\documentclass{article}

\usepackage{arxiv}

\usepackage[utf8]{inputenc} 
\usepackage[T1]{fontenc}    
\usepackage{hyperref}       
\usepackage{url}            
\usepackage{booktabs}       
\usepackage{amsfonts}       
\usepackage{nicefrac}       
\usepackage{microtype}      
\usepackage{lipsum}
\usepackage{graphicx}
\usepackage{acronym}
\usepackage{multirow}
\usepackage{adjustbox}
\usepackage{diagbox}
\usepackage{amsmath}
\usepackage{subfigure}
\usepackage{algorithm}
\usepackage{algorithmic}
\usepackage[numbers]{natbib}

\title{HADL Framework for Noise Resilient Long-Term Time Series Forecasting}

\author{
 Aditya Dey \\
  Department of Data Science\\
  Norwegian University of Life Sciences (NMBU)\\
  Aas, Norway 1432 \\
  \texttt{aditya.dey@nmbu.no} \\
   \And
 Jonas Kusch \\
  Department of Data Science\\
  Norwegian University of Life Sciences (NMBU)\\
  Aas, Norway 1432 \\
  \texttt{jonas.kusch@nmbu.no} \\
  \And
 Fadi Al Machot \\
  Department of Data Science\\
  Norwegian University of Life Sciences (NMBU)\\
  Aas, Norway 1432 \\
  \texttt{fadi.al.machot@nmbu.no} \\
}

\begin{document}
\acrodef{AI}[AI]{Artificial Intelligence}
\acrodef{ML}[ML]{Machine Learning}
\acrodef{DL}[DL]{Deep Learning}
\acrodef{RNN}[RNN]{Recurrent Neural Network}
\acrodef{LSTM}[LSTM]{Long Short-Term Memory}
\acrodef{GRU}[GRU]{Gated Recurrent Unit}
\acrodef{LLM}[LLM]{Large Language Models}
\acrodef{NLP}[NLP]{Natural Language Processing}
\acrodef{ARIMA}[ARIMA]{Autoregressive Integrated Moving Average}
\acrodef{SARIMA}[SARIMA]{Seasonal ARIMA}
\acrodef{LTSF}[LTSF]{Long-term Time Series Forecasting}
\acrodef{LLM}[LLM]{Large Language Models}
\acrodef{CeNN}[CeNN]{Cellular Neural Network}
\acrodef{ESN}[ESN]{Echo State Network}
\acrodef{SOH}[SOH]{State of Health}
\acrodef{XAI}[XAI]{Explainable Artificial Intelligence.}
\acrodef{VAR}[VAR]{Vector Auto Regressive}
\acrodef{ANN}[ANN]{Artificial Neural Network}
\acrodef{MORAI}[MORAI]{Masked Encoder-based Universal Time Series Forecasting Transformer}
\acrodef{LOTSA}[LOTSA]{Large-scale Open Time Series Archive}
\acrodef{CV}[CV]{Computer Vision}
\acrodef{CNN}[CNN]{Convolutional Neural Network}
\acrodef{TCN}[TCN]{Temporal Convolution Neural Network}
\acrodef{MLP}[MLP]{Multi-layer Perceptron}
\acrodef{ESN}[ESN]{Echo State Network}
\acrodef{SOTA}[SOTA]{State of the Art}
\acrodef{TSF}[TSF]{Time Series Forecasting}
\acrodef{DFT}[DFT]{Discrete Fourier Transform}
\acrodef{DWT}[DWT]{Discrete Wavelet Transform}
\acrodef{DCT}[DCT]{Discrete Cosine Transform}
\acrodef{DST}[DST]{Discrete Sine Transform}
\acrodef{STFT}[STFT]{Short Time Fourier Transform}
\acrodef{MSE}[MSE]{Mean Squared Error}
\acrodef{MAE}[MAE]{Mean Absolute Error}
\acrodef{NRR}[NRR]{Noise-Resilience Relative}
\acrodef{MAV}[MAV]{Mean Absolute Variability}

\maketitle
\begin{abstract}
Long-term time series forecasting is critical in domains such as finance, economics, and energy, where accurate and reliable predictions over extended horizons drive strategic decision-making. Despite the progress in machine learning-based models, the impact of temporal noise in extended lookback windows remains underexplored, often degrading model performance and computational efficiency. In this paper, we propose a novel framework that addresses these challenges by integrating the Discrete Wavelet Transform (DWT) and Discrete Cosine Transform (DCT) to perform noise reduction and extract robust long-term features. These transformations enable the separation of meaningful temporal patterns from noise in both the time and frequency domains. To complement this, we introduce a lightweight low-rank linear prediction layer that not only reduces the influence of residual noise but also improves memory efficiency. Our approach demonstrates competitive robustness to noisy input, significantly reduces computational complexity, and achieves competitive or state-of-the-art forecasting performance across diverse benchmark datasets. Extensive experiments reveal that the proposed framework is particularly effective in scenarios with high noise levels or irregular patterns, making it well suited for real-world forecasting tasks. The code is available in \url{https://github.com/forgee-master/HADL}.
\end{abstract}


\section{Introduction} \label{sec: intro}
Real-world time series data from domains such as finance \cite{idrees2019prediction}, economics \cite{li2023relationship}, and energy systems \cite{deb2017review} rely on accurate long-term forecasting to support informed decision-making that has a significant impact on global development \cite{yaslan2017empirical}. However, these datasets often contain significant levels of temporal noise due to their highly volatile nature \cite{cornell2024probabilistic}, environmental factors \cite{gandhmal2019systematic}, and random fluctuations. This poses a critical challenge for \ac{LTSF} models, as they can obscure the underlying patterns and reduce the accuracy of the predictions \cite{zeng2023transformers}. Thus, a systematic study of temporal noise in \ac{LTSF} and its effects on model accuracy is required. 

Most \ac{LTSF} transformer models can perform well with a small lookback window due to the large parameter size and the benefits of using vanilla or modified attention \cite{nie2022time, liu2023itransformer,zhang2024frnet}. However, the gains in accuracy from increasing lookback windows are not significant \cite{nie2022time}, which may be due to temporal noise that obscures the underlying trend and seasonality patterns. This highlights the trade-off between a large lookback window, parameter size, and temporal noise, where increasing the lookback window or parameter size may not yield significant accuracy gains due to noise.

Although these large-parameter models perform well with a small lookback window, low-parameter \ac{MLP} models with a larger lookback window have matched or outperformed them while using fewer parameters and lower computational complexity \cite{xu2023fits,lin2024sparsetsf}. An analysis of these models suggests that underlying temporal patterns in time series can be extracted using simple attention \cite{ekambaram2023tsmixer}, trend-seasonality decomposition \cite{zeng2023transformers,lin2024sparsetsf}, or time-to-frequency domain conversions \cite{xu2023fits}. The success of these low-parameter models challenges the conventional reliance on large parameters, demonstrating that accurate \ac{LTSF} is achievable with streamlined architectures. 

These low-parameter models, though computationally efficient, also face a trade-off between lookback window size and temporal noise. A small lookback window may not provide sufficient information, and a larger window is susceptible to temporal noise \cite{zeng2023transformers}. Hence, these models are based on a sufficiently large lookback to achieve good performance \cite{xu2023fits,lin2024sparsetsf}. 

Thus, to address the challenges of noise robustness, lightweight design, and accuracy improvement in \ac{LTSF}, we propose the HADL framework. This framework employs \ac{DWT} with a Haar wavelet to reduce the effects of noise in a large lookback window, effectively compressing the input length by half through approximation coefficients. The compressed series is then transformed into the frequency domain using \ac{DCT}, enabling high-resolution extraction of underlying patterns. Finally, a low-rank layer improves generalization while maintaining minimal memory requirements. We avoid channel mixing, which can accidentally project the noise from one series onto another \cite{nie2022time} and utilize a general layer instead of a per channel layer, which helps to reduce the risk of overfitting by sharing parameters across channels. 

The HADL framework integrates \ac{DWT}, \ac{DCT}, and a low-rank layer, offering the following advantages:
\begin{itemize}
\item \ac{DWT} reduces noise and compresses the input, enabling a lightweight model with fewer parameters.
\item \ac{DCT} enhances the extraction of long-term temporal patterns while reducing computational overhead. 
\item Low-rank layers improve generalization and noise robustness while further minimizing parameter requirements.
\item A single prediction layer reduces overall complexity, ensuring that the model remains lightweight and interpretable.
\end{itemize}
This framework achieves accuracy comparable to or better than other \ac{SOTA} models while significantly reducing parameter size and effectively mitigating noise. The result is a lightweight, robust model specifically designed to perform well on real-world noisy datasets with highest accuracy.

\section{Related Works} \label{sec: related works}

\subsection{MLP-Based Models for Lightweight Forecasting} \label{sub sec: related works MLP}
\ac{MLP} models offer computationally efficient and interpretable alternatives, often achieving predictive performance that is comparable to or superior to other approaches \cite{zeng2023transformers,xu2023fits,ekambaram2023tsmixer}. For example, DLinear \cite{zeng2023transformers} applies trend and seasonality decomposition, whereas FITS \cite{xu2023fits} achieves higher accuracy using \ac{DFT} with approximately $5$K parameters. Similarly, SparseTSF \cite{lin2024sparsetsf}, an ultra-lightweight model with approximately $1$K parameters, effectively utilizes cross-period sparse techniques to extract periodic features. Other notable \ac{MLP}-based models, such as TSMixer \cite{ekambaram2023tsmixer}, LightCTS \cite{lai2023lightcts}, and FreTS \cite{yi2024frequency}, may not be as lightweight as SparseTSF but nevertheless deliver competitive performance. 

Some of these models rely on techniques that extract periodicity in the time domain by selecting the period length as a hyperparameter \cite{lin2024sparsetsf,yi2024frequency} or by converting the data into the frequency domain using the \ac{DFT} \cite{xu2023fits,yi2024frequency}. Time domain approaches often lack robust methods for reducing temporal noise, and the assumption that all features share the same periodicity set as a hyperparameter can lead to inaccuracies \cite{lin2024sparsetsf}. On the other hand, frequency-domain methods using \ac{DFT} with cutoff frequencies to reduce temporal noise assume a uniform cutoff range for all features, which may impact accuracy.

\subsection{Convolutional and Recurrent Neural Network Methods} \label{sub sec: related works CNN}

Other classes of models, such as ModernTCN \cite{luo2024moderntcn}, employ dilated temporal convolutions to extract and enhance temporal characteristics, reaffirming the competitiveness of \ac{CNN}s. Although not as lightweight as \ac{MLP} models, ModernTCN demonstrates comparable runtime efficiency. Similarly, \ac{RNN}-based models, such as SegRNN \cite{lin2023segrnn}, xLSTM \cite{alharthi2024xlstmtime}, and WiTRAN \cite{jia2024witran}, excel in capturing temporal dependencies and sequence relationships. However, like other approaches, these models rely on appropriate lookback window sizes and large parameter counts, which may be affected by temporal noise.


\subsection{Transformer-based Models} \label{sub sec: related works transformers}
Most emerging \ac{LTSF} models are transformer-based \cite{liu2024deep}, with notable examples such as PatchTST, which demonstrates that modifications to the attention mechanism can be avoided by segmenting univariate time series into patches \cite{nie2022time}. The iTransformer extends this concept further by treating each feature as a patch, allowing a receptive field that spans the entire length of the feature \cite{liu2023itransformer}. Although these patching techniques can enhance model performance, they also introduce challenges. For example, a noisy patch can obscure the underlying patterns. Similarly, in iTransformer, noisy features may impede the model's ability to discern meaningful structures. These limitations underscore the sensitivity of patch-based methods to temporal noise, posing challenges to consistent performance.

Despite these challenges, transformer-based models continue to inspire advancement, with recent models, such as FredFormer \cite{piao2024fredformer} and FrNet \cite{zhang2024frnet}, aiming to further enhance prediction accuracy. However, these transformers often require significant memory and computational resources, making them less practical for scenarios where efficiency is critical.

Low-rank implementations have not been widely studied in \ac{LTSF} models. However, they have shown potential in vanilla transformers by improving generalization \cite{hu2021lora,zhang2023adalora,valipour2022dylora}. They have also been shown to enhance inter-channel dependencies in \ac{LTSF} \cite{nie2024channel}. Incorporating low-rank techniques can improve performance in time series forecasting by efficiently capturing critical temporal patterns while reducing parameter complexity.

Currently, no \ac{LTSF} models have simultaneously addressed noise robustness, accuracy improvement, and lightweight design in a unified manner.

\section{The Proposed Method} \label{sec: proposed method}

\begin{figure*}[htbp]
    \centering
    \includegraphics[width=\textwidth]{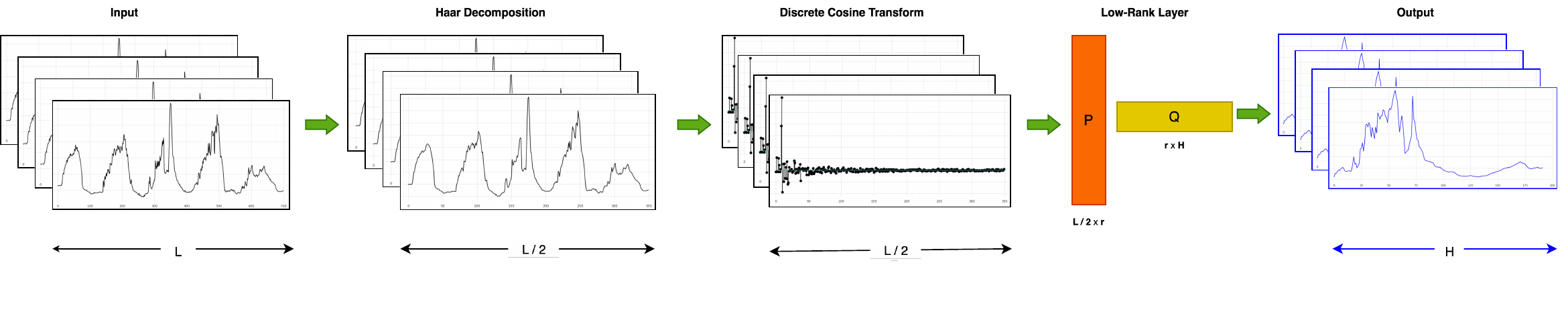}
    \caption{Illustration of the HADL architecture.}
    \label{fig:haar-dct-model}
\end{figure*}

In this section, we provide a concise overview of the HADL framework, followed by a detailed explanation and rationale for the design choices.

\textbf{Problem Definition:} Given a regularly sampled multivariate time series with a lookback window $L$, represented as $X_T = {x_1, \dots, x_t, \dots, x_L} \in \mathbb{R}^{C \times L}$, where each time step $x_t$ is a vector of dimension $C$, the objective is to forecast future values $\hat{Y}_T = {x_{L+1}, \dots, x_{L+\tau}, \dots, x_{L+H}} \in \mathbb{R}^{C \times H}$ over a horizon of length $H$.

\textbf{Summary:} The multivariate time series $X_T \in \mathbb{R}^{C \times L}$ is first transformed by \ac{DWT} using Haar wavelet to obtain approximation coefficient values $A_T \in \mathbb{R}^{C \times \frac{L}{2}}$, effectively compressing the input length by half while removing the detail coefficient values to mitigate the impact of noise. Next, the time domain representation $A_T$ is converted into the frequency domain $A_F \in \mathbb{R}^{C \times \frac{L}{2}}$ using the \ac{DCT}, allowing the model to extract long-term periodic features at high-frequency resolution. Finally, the frequency representation $A_F$ is processed through a low-rank layer to predict the target $\mathbf{\hat{Y}} \in \mathbb{R}^{C \times H}$ in the time domain, eliminating the need for an explicit inverse transform. 
\begin{subequations} \label{equation: HADL equations}
\begin{align}
    A_T, \_ &= \text{DWT}(X_T) \label{eq:haar-decomposition} \\
    A_F &= \frac{2}{L} \times \text{DCT}(A_T) \label{eq:discrete-cosine-transform} \\
    \hat{Y}_T &= \text{LowRankLayer}(A_F) \label{eq:low-rank-prediction}
\end{align}
\end{subequations}
The complete architecture of the HADL framework is illustrated in Figure~\ref{fig:haar-dct-model}. The transformation and prediction processes are formally described by the equations in \eqref{equation: HADL equations} and the algorithm presented in Appendix \ref{appendix: algorithm}.

\subsection{Discrete Wavelet Transform} \label{sec: Discrete Wavelet Transform}
The trend-seasonality decomposition in the time domain provides high temporal resolution but relies heavily on accurately selecting the seasonal window to capture periodic patterns \cite{zeng2023transformers,lin2024sparsetsf}. An incorrect choice of window size can lead to temporal leakage or distortion of periodic components. In contrast, the \ac{DFT} offers high-frequency resolution \cite{xu2023fits,yi2024frequency} but operates under the implicit assumption that the observed time series represents a complete cycle of a single period \cite{ahmed1974discrete}. Violating this assumption results in spectral leakage, which distorts the frequency components. A balanced resolution in both the time and frequency domains can be achieved through methods like \ac{STFT} or \ac{DWT}, which provide multiresolution decomposition at the cost of increased dimensionality. 

The key advantage of \ac{DWT} over \ac{STFT} lies in the orthogonality of wavelet functions, which ensures that at each level of decomposition, distinct frequency components are captured \cite{daubechies1993ten}. This property eliminates redundancy and improves the interpretability of transformed data, addressing the main limitation of \ac{STFT}, where overlapping windowed components can redundantly capture the same frequency across multiple levels.

Therefore, our method begins with \ac{DWT} to decompose the time series. Among various wavelet families, we choose the Haar wavelet because of its simplicity and interpretability. The Haar wavelet performs convolution on the time series with two sets of filters: an approximation filter and a detail filter with a kernel size of $2$ and a stride of $2$. The approximation filter is an averaging operator, while the detail filter acts as a differencing operator. For our purposes, we use the approximation coefficients $A_T \in \mathbb{R}^{C \times \frac{L}{2}}$, which capture essential information while reducing noise. The detail coefficients $D_T \in \mathbb{R}^{C \times \frac{L}{2}}$, which mainly represent noise, are discarded. 
\begin{align}
A_T &= X_T * \left[ \frac{1}{\sqrt{2}}, \frac{1}{\sqrt{2}} \right] \label{eq:approximation-filter}\\
D_T &= X_T * \left[ -\frac{1}{\sqrt{2}}, \frac{1}{\sqrt{2}} \right] \label{eq:detail-filter}
\end{align}
Although it is possible to use detail coefficients for prediction with linear layers, doing so would increase the model's complexity and parameter count without substantial benefits.

We limit the decomposition to a single level to avoid excessive smoothing, which can eliminate critical frequency components necessary for accurate forecasting. For lookback windows in the typical range of $336$ to $720$, a single level of decomposition is sufficient. However, multiple decomposition levels may be considered for ultra-long lookback windows. Additionally, employing linear layers across multiple decomposition levels significantly increases computational complexity and may not provide substantial improvements in feature extraction.

\subsection{Discrete Cosine Transform}

\ac{DFT} and \ac{DCT} share a common mathematical foundation but differ in their focus and application. \ac{DFT} represents a signal using both real and imaginary components. The cosine component captures smooth, low-frequency variations, while the sine component highlights abrupt changes and high-frequency components \cite{ahmed1974discrete}. Although this dual representation is comprehensive, the \ac{DCT}'s exclusive focus on cosine components leads to energy compaction, in which most of the signal's energy is concentrated in fewer coefficients \cite{soon1998noisy,baharanchi2022noise}. This makes \ac{DCT} particularly effective in representing long-term patterns in \ac{LTSF} while avoiding the additional complexity introduced by imaginary parts.

Thus, the approximation coefficients $A_T \in \mathbb{R}^{C \times \frac{L}{2}}$ undergo \ac{DCT} Type II conversion, generating an output $A_F \in \mathbb{R}^{C \times \frac{L}{2}}$ with high-frequency resolution. This transformation is accompanied by length-based normalization ($\frac{1}{\frac{L}{2}}$), ensuring that the resulting frequency domain representation is scale-independent and numerically stable.
\begin{equation} \label{eq: discrete cosine transformation}
A_F = \frac{2}{L} \cdot \text{DCT}(A_T)
\end{equation}
When \ac{DCT} is applied after \ac{DWT} instead of directly on the original time series, the process offers distinct advantages. The approximation coefficients from \ac{DWT} provide a smoothed, noise-reduced representation of the data. This preprocessing ensures that the \ac{DCT} operates on a cleaner input, avoiding distortions caused by high-frequency noise in the original series. Additionally, reducing the series length decreases the computational complexity of the \ac{DCT} operation and focuses the analysis on the most informative components of the signal, ensuring that critical long-term periodic features are emphasized.

This combination is advantageous in real-world forecasting as it balances noise robustness, computational efficiency, and the extraction of long-term dependencies and periodic structures.

\subsection{Low Rank Layer}\label{subsec: Low Rank layer}

Most models in \ac{LTSF} employ standard linear layers for the final prediction; however, using a low-rank layer offers several advantages. Low-rank layers have a significantly reduced memory footprint during training and inference. Moreover, a reduced memory footprint directly leads to fewer operations required to train and evaluate the network.


There are different approaches to defining a low-rank layer. Perhaps the most common strategy for defining a low-rank layer is selecting two matrices: $P \in \mathbb{R}^{\frac{L}{2} \times r}$ and $Q \in \mathbb{R}^{r \times H}$, where $r$ represents the rank of the layer ($r \ll \min(\frac{L}{2}, H)$) \cite{hu2021lora}. These matrices jointly form a low-rank approximation of the weight matrix $W\approx PQ$, which significantly reduces the number of trainable parameters. In addition, a bias vector $B \in \mathbb{R}^H$ is included to account for linear shifts. The low-rank layer then takes the form:
\begin{equation}\label{eq: low-rank equation}
    \hat{Y}_T = A_F P Q + B\,.
\end{equation}
Standard gradient-based optimization methods are used to train $P$, $Q$, and $B$. The corresponding gradient flow for a loss function $\mathcal{L}(PQ)$ (where we omit dependence on the bias and $A_F$ for ease of presentation) then reads
\begin{subequations}\label{eq:PQevolution}
    \begin{align}
    \dot{P}(t) =\,& -\nabla_{P} \mathcal{L}(P(t)Q(t))\,, \\
    \dot{Q}(t) =\,& -\nabla_{Q} \mathcal{L}(P(t)Q(t))\,.
\end{align}
\end{subequations}
 Here, $\nabla_{P,Q}$ denotes the gradients with respect to the respective low-rank factors. If the method converges to a stationary point of \eqref{eq:PQevolution} at time $t_{\star}$ and $W(t_{\star}) := P(t_{\star})Q(t_{\star})$ satisfies the gradient flow equation $\dot{W} = -\nabla_{W} \mathcal{L}(W)$ at time $t_{\star}$, then $\nabla_{W}\mathcal{L}(W(t_{\star})) = 0$ since
\begin{align*}
    \nabla_{W}\mathcal{L}(W(t_{\star})) =\,& -\dot{W}(t_{\star})\\
    =\,& -\dot{P}(t_{\star})Q(t_{\star}) - P(t_{\star})\dot{Q}(t_{\star}) = 0\,.
\end{align*}
In this case, the stationary point fulfills the necessary condition of a local full-rank optimum. A primary advantage of low-rank layers is their computational efficiency, which results from the fewer operations required to compute \eqref{eq: low-rank equation}. This lower computational cost in the forward pass directly translates into fewer evaluations in the backward pass, leading to shorter training time. In addition to computational efficiency, low-rank networks have been observed to outperform their dense counterparts \cite{lee2019snip,schotthofer2022low}. This observation is not yet rigorously understood, but it is assumed that the regularization imposed by low-rank compression improves the generalization of the resulting networks. In Section \ref{sub sec: abalation low rank layer}, we study the regularization effects of low-rank networks and show that similar behavior can be observed in our experiments.

\subsection{Key Features} \label{sub sec: key features}

A distinctive feature of our framework is the omission of the inverse transformation from the frequency domain to the time domain. This simplification is justified by Parseval's theorem, as shown in Equation \eqref{eq: parseval theorem}, which ensures energy conservation between the time and frequency domains \cite{yi2024frequency}.

\begin{equation} 
\label{eq: parseval theorem} \sum \limits_{n=0}^{N-1}|(A_T)_n|^2 = \frac{1}{N} \sum \limits_{n=0}^{N-1} |(A_F)_n|^2 
\end{equation}
As a result, the model operates directly in the frequency domain, capturing necessary relationships without the added complexity of domain conversion. This design choice enhances computational efficiency and interpretability while preserving the fidelity of the original time series.

Additionally, the framework employs a single generalized low-rank layer instead of separate layers for each feature or channel. This design reduces parameter redundancy, mitigates overfitting to specific channels, and enhances overall model efficiency. With the low-rank layer as the only trainable component, the model achieves a compact architecture that balances simplicity with effective noise mitigation. This streamlined approach makes the HADL framework particularly suitable for \ac{LTSF} tasks, offering both accuracy and computational efficiency.

To further mitigate overfitting and reduce noise influence, L1 regularization is incorporated. This regularization shrinks the coefficients of less important features, contributing to a more robust and generalizable model.

\section{Experiments}\label{sec: experiments}

We conduct a standard \ac{LTSF} test for prediction lengths \( H = \{96, 192, 336, 720\} \), using a fixed lookback window of \( L = 512 \) for all baseline models.

Additionally, we conduct a robustness test by introducing noise into the training data (\( X_{\text{train}}, y_{\text{train}} \)) as follows:
\begin{equation}
X_{\text{train}} := X_{\text{train}} + \mathcal{N}(0,1) \cdot \eta
\end{equation}
Here, \( \mathcal{N}(0,1) \) represents random noise sampled from a standard normal distribution, and \( \eta = \{0.0, 0.3, 0.7, 1.7, 2.3\} \) controls the noise intensity. The validation and test datasets remain unchanged to ensure consistent evaluation of the model's robustness.

\textbf{Settings} All experiments were conducted using the PyTorch \cite{imambi2021pytorch} and SciPy \cite{2020SciPy-NMeth} frameworks on an NVIDIA Quadro RTX 8000 GPU with 100 GB of memory. The models were trained using the ADAM optimizer for 100 epochs with a patience of 20 epochs for multivariate forecasting and 50 epochs with a patience of 10 epochs for robustness testing.

\textbf{Datasets} We evaluate our model on four widely used benchmark datasets: ETT \cite{zhou2021informer} (ETTh1, ETTh2, ETTm1, ETTm2) and Traffic \cite{lai2018modeling}. Detailed descriptions of these datasets are provided in Appendix \ref{appendix: Datasets}.

\textbf{Baseline Models} We compare our approach against several recent and highly successful \ac{SOTA} models, including PatchTST \cite{nie2022time}, iTransformer \cite{liu2023itransformer}, FrNet \cite{zhang2024frnet}, DLinear \cite{zeng2023transformers}, SparseTSF \cite{lin2024sparsetsf}, FreTS \cite{yi2024frequency}, and ModernTCN \cite{luo2024moderntcn}.

\textbf{Evaluation Metrics} Prediction errors for multivariate forecasting and robustness testing are evaluated using \ac{MSE} and \ac{MAE}.

For multivariate forecasting, we report Improvement (Imp.), which quantifies the error reduction achieved by our model compared to the best-performing baseline model. This is calculated as the difference between the \ac{MSE} of the best baseline model ($\text{MSE}_{\text{baseline}}$) and the \ac{MSE} of our model ($\text{MSE}_{\text{ours}}$). A higher positive improvement value indicates that our model performs significantly better than the baseline, while a lower or negative value suggests comparable or worse performance.
\begin{equation}
\text{Imp.} = \text{MSE}_{\text{baseline}} - \text{MSE}_{\text{ours}}
\end{equation}
For robustness testing, we introduce the concept of \ac{NRR} to better evaluate the impact of increasing noise intensity. \ac{NRR} quantifies the deviation in prediction error as the noise intensity increases, offering a clear perspective on the model's robustness.
\begin{align}
  \text{NRR} = \frac{\text{MSE}_{\eta = x}}{\text{MSE}_{\eta = 0.0}} \\
\text{MAV} = \frac{1}{n} \sum_{i=1}^{n} |\text{NRR}_i - 1|
\end{align}  
This metric allows us to compare the model's performance under noisy conditions ($\eta=x$) with its baseline performance ($\eta=0.0$). To summarize the overall performance of the model across varying noise intensities, we compute the \ac{MAV} of the \ac{NRR} with central point of $1$. A lower \ac{MAV} indicates that the model maintains consistent performance as the noise increases, demonstrating robustness.

\section{Results}

In this section, we discuss Multivariate Forecasting and Robustness Testing, followed by a combined analysis.

\begin{table*}[hbt!]
    \centering
    
    \caption{Results for MSE (lower is better) in long-term time series forecasting on benchmark datasets. We compare HADL framework with (w) and without (w/o) regularization against baseline models using standard prediction lengths $H = \{96, 192, 336, 720\}$ and a lookback window $L=512$ for all models. The best results are in \textbf{bold}, and the second best are \underline{underlined}. "Imp." denotes the improvement compared to best-performing baseline model.}
    \begin{adjustbox}{width=\textwidth}
    \renewcommand{\arraystretch}{1.5}
    \begin{tabular}{c|c|cccc|cccc|cccc|cccc|cccc} 
   
        \multirow{2}{*}{Model Type} & Datasets & \multicolumn{4}{c|}{ETTh1} & \multicolumn{4}{c|}{ETTh2} & \multicolumn{4}{c|}{ETTm1} & \multicolumn{4}{c|}{ETTm2} & \multicolumn{4}{c}{Traffic} \\
       & H & 96 & 192 & 336 & 720 & 96 & 192 & 336 & 720 & 96 & 192 & 336 & 720 & 96 & 192 & 336 & 720 & 96 & 192 & 336 & 720  \\
       \hline
       \multirow{4}{*}{Transformer}& PatchTST (\citeyear{nie2022time}) & 0.394 & 0.424 & 0.457 & 0.461 & 0.292 & 0.347 & 0.380 & 0.431 & \underline{0.293} & \underline{0.335}  & 0.368 & 0.423 & \underline{0.164} & 0.223 & 0.277 & 0.381 & 0.489 & 0.498 & 0.506 & 0.541 \\
       & iTransformer (\citeyear{liu2023itransformer}) & 0.399 & 0.428 & 0.472 & 0.685 & 0.301 & 0.381 & 0.395 & 0.467 & 0.315 & 0.350 & 0394 & 0.449 & 0.188 & 0.250 & 0.326 & 0.391 & 0.433 & 0.462 & 0.443 & 0.489 \\
       & FrNet (\citeyear{zhang2024frnet}) & \textbf{0.362} & \textbf{0.390} & 0.427 & 0.468 & \textbf{0.269} & \textbf{0.330} & \textbf{0.358} & \underline{0.391} & 0.295 & 0.337 & \textbf{0.363} & \underline{0.421} & 0.165 & \underline{0.221} & \underline{0.274} & \underline{0.360} & \textbf{0.367} & \textbf{0.390} & \textbf{0.397} & \textbf{0.433} \\
       \hline
       CNN & ModernTCN (\citeyear{luo2024moderntcn}) & 0.472 & 0.498 & 0.507 & 0.531 & 0.300 & 0.348 & 0.384 & 0.410 & 0.313 & 0.363 & 0.392 & 0.447 & 0.169 & 0.232 & 0.307 & 0.385 & \underline{0.392} & \underline{0.406} & \underline{0.415} & \underline{0.464} \\
       \hline
       \multirow{5}{*}{MLP}& DLinear (\citeyear{zeng2023transformers}) & 0.385 & 0.422 & 0.457 & 0.496 & 0.390 & 0.512 & 0.697 & 1.060 & \textbf{0.292} & \textbf{0.333} & 0.368 & 0.428 & 0.204 & 0.333 & 0.342 & 0.570 & 0.418 & 0.430 & 0.440 & 0.479\\
       & SparseTSF (\citeyear{lin2024sparsetsf}) & 0.396 & 0.412 & \underline{0.426} & \textbf{0.427} & 0.285 & 0.345 & 0.364 & \textbf{0.384} & 0.330 & 0.347 & 0.374 & 0.424  & 0.173 & 0.225 & 0.277 & 0.362& 0.427 & 0.441 & 0.454 & 0.487\\
       & FreTS (\citeyear{yi2024frequency}) & 0.396 & 0.435 & 0.471 & 0.565 & 0.318 & 0.432 & 0.479 & 0.809 & 0.317 & 0.356 & 0.385 & 0.440 & 0.179 & 0.254 & 0.314 & 0.389 & 0.398 & 0.414 & 0.428 & 0.471 \\
       \cline{2-22}
       & HADL (w/o) & 0.365 & 0.397 & 0.423 & 0.446 & 0.272 & 0.334 & 0.358 & 0.395 & 0.306 & 0.339 & 0.367 & 0.420 & 0.163 & 0.218 & 0.271 & 0.359 & 0.413 & 0.433 & 0.444 & 0.481\\
       & HADL (w/) & \underline{0.363} & \underline{0.395} & \textbf{0.421} & \underline{0.444} & \underline{0.271} & \underline{0.334} & \textbf{0.358} & 0.395 & 0.304 & 0.337 & \underline{0.365} & \textbf{0.418} & \textbf{0.163} & \textbf{0.218} & \textbf{0.271} & \textbf{0.359} & 0.412 & 0.433 & 0.445 & 0.481\\
       \hline
       \hline
       \multicolumn{2}{c}{Imp.} & -0.001&-0.005& +0.005&-0.017&-0.002&-0.004&0.000&-0.011&-0.012&-0.005&-0.002&+0.003&+0.001&+0.003&+0.003&+0.001&-0.045&-0.043&-0.048&-0.048
        \label{tab: Multivariate Forecasting}
    \end{tabular}
    \end{adjustbox}
    

    \centering
    \caption{Results comparing FLOPs and total trainable parameters between HADL framework against baseline models on the ETTh1 and Traffic datasets for multivariate forecasting, using standard prediction lengths $H = \{96, 192, 336, 720\}$ and a lookback window $L=512$ for all models.}
    \begin{adjustbox}{width=\textwidth}
    \renewcommand{\arraystretch}{1.5}
    \begin{tabular}{c|c|cccc|cccc||cccc|cccc} 
    & &  \multicolumn{8}{c||}{FLOPs ($\times 10^9$)} & \multicolumn{8}{c}{Parameters ($\times 10^3$)} \\
    \cline{3-18}
    \multirow{2}{*}{Model Type} & Dataset & \multicolumn{4}{c|}{ETTh1} & \multicolumn{4}{c||}{Traffic} & \multicolumn{4}{c|}{ETTh1} & \multicolumn{4}{c}{Traffic} \\
       & H & 96 & 192 & 336 & 720 & 96 & 192 & 336 & 720 & 96 & 192 & 336 & 720 & 96 & 192 & 336 & 720 \\
       \hline
       \multirow{3}{*}{Transformers} 
       & PatchTST (\citeyear{nie2022time}) & 0.336 & 0.358 & 0.391 & 0.479 & 41.465 & 44.176 & 48.244 & 59.090 & 706.2 & 1395.0 & 2428.2 & 5183.4 & 84838.2 & 169659.0 & 296890.2 & 636173.4 \\
       & iTransformer (\citeyear{liu2023itransformer}) & 0.337 & 0.346 & 1.272 & 1.341 & 50.864 & 51.545 & 148.701 & 154.149 & 948.0 & 972.7 & 3592.0 & 3789.0 & 948.0 & 972.7 & 3592.0 & 3789.0 \\
       & FrNet (\citeyear{zhang2024frnet}) & 0.324 & 0.252 & 0.538 & 5.659 & 1.285 & 1.293 & 1.308 & 1.351 & 116.5 & 196.9 & 577.6 & 7485.9 & 81950.8 & 82614.8 & 83714.3 & 87254.8 \\
       \hline
       CNN 
       & ModernTCN (\citeyear{luo2024moderntcn}) & 0.568 & 0.745 & 1.009 & 1.713 & 456.455 & 478.148 & 510.687 & 597.459 & 930.1 & 1716.7 & 2896.5 & 6042.6 & 10639.7 & 107326.3 & 108506.1 & 111652.2 \\
       \hline
       \multirow{4}{*}{MLP} 
       & DLinear (\citeyear{zeng2023transformers}) & 0.022 & 0.044 & 0.077 & 0.165 & 2.711 & 5.423 & 9.490 & 20.3337 & 1034.2 & 2068.4 & 3619.7 & 7756.5 & 127355.3 & 254710.6 & 445743.6 & 955164.9 \\
       & SparseTSF (\citeyear{lin2024sparsetsf}) & 0.002 & 0.003 & 0.004 & 0.007 & 0.324 & 0.409 & 0.536 & 0.875 & 0.2 & 0.4 & 0.6 & 1.4 & 0.2 & 0.4 & 0.6 & 1.4 \\
       & FreTS (\citeyear{yi2024frequency}) & 3.778 & 3.783 & 3.792 & 3.814 & 14.539 & 14.561 & 14.592 & 14.677 & 16868.3 & 16892.9 & 16930.0 & 17028.6 & 16868.3 & 16892.9 & 16930.0 & 17028.6 \\
          \cline{2-18}
       & HADL & 0.004 & 0.005 & 0.006 & 0.011 & 0.499 & 0.632 & 0.830 & 1.360 & 17.6 & 22.5 & 29.9 & 49.5 & 17.6 & 22.5 & 29.9 & 49.5
       \label{tab: params and flops multivariate forecasting}
    \end{tabular}
    \end{adjustbox}
\end{table*}

\textbf{Multivariate Forecasting} Table \ref{tab: Multivariate Forecasting} highlights HADL’s strengths in long-term multivariate forecasting in various datasets. It achieves competitive or superior performance compared to baseline models while maintaining significantly fewer trainable parameters and lower memory consumption (Table \ref{tab: params and flops multivariate forecasting}). The complete benchmark results, evaluated using \ac{MSE} and \ac{MAE}, are provided in Appendix \ref{appendix sub: Multivariate Forecasting results}.

HADL ranks among the best performers, often achieving the best or second-best \ac{MSE}. In ETTh2, it improves \ac{MSE} by $+0.001$ to $+0.003$ across all horizons while using only $17$--$50$K parameters, far fewer than FrNet ($116$--$7486$K). Across ETTh1, ETTh2 and ETTm1, it achieves mixed results ($-0.001$ to $+0.012$), maintaining efficiency over DLinear ($1034$--$7756$K) and PatchTST ($706$--$5184$K). Its lightweight design results in lower FLOPs ($0.004$--$0.011$B).

In the Traffic dataset, HADL struggles in high-dimensional settings, showing negative improvements ($-0.045$ to $-0.048$), where transformer-based models excel. However, its performance against DLinear remains close ($0.002$--$0.006$), indicating the effectiveness of its generalization approach in \ac{MLP} based models. Notably, HADL outperforms SparseTSF, which, despite having the smallest parameter count ($0.2$–$1.4$K), fails to leverage minimal parameters effectively.

\begin{table*}[h]
    \centering
    \caption{Noise-Resilience Relative (NRR) and Mean Absolute (MAV) results for robustness testing on benchmark datasets, comparing HADL framework with (w/) and without (w/o) regularization against baseline models. The tests are conducted using a lookback window of $L=512$ and a prediction length of $H=192$ under varying noise intensities $\eta = \{0.0, 0.3, 0.7, 1.3, 1.7, 2.3\}$. The best results are in \textbf{bold}, and the second best are \underline{underlined}.}
    \begin{adjustbox}{width=\textwidth}
    \renewcommand{\arraystretch}{2.0}
    \begin{tabular}{c|c|cccc|c|cccc|c|cccc|c|cccc|c|cccc|c} 
     &  & \multicolumn{5}{c|}{ETTh1} & \multicolumn{5}{c|}{ETTh2} & \multicolumn{5}{c|}{ETTm1} & \multicolumn{5}{c|}{ETTm2} & \multicolumn{5}{c}{Traffic} \\
    \cline{3-27}
    \multirow{2}{*}{Model Type} & $\eta$ & 0.3 & 0.7 & 1.3 & 1.7 &  & 0.3 & 0.7 & 1.3 & 1.7 & & 0.3 & 0.7 & 1.3 & 1.7 &  & 0.3 & 0.7 & 1.3 & 1.7 &  & 0.3 & 0.7 & 1.3 & 1.7 &  \\
    & & NRR & NRR & NRR & NRR & MAV & NRR & NRR & NRR & NRR & MAV & NRR & NRR & NRR & NRR & MAV & NRR & NRR & NRR & NRR & MAV & NRR & NRR & NRR & NRR & MAV \\
    \hline
    \hline
    \multirow{3}{*}{Transformers} 
       & PatchTST (\citeyear{nie2022time}) & 1.002 & 1.007 & 1.028 & 1.044 & 0.020 & 1.008 & 1.011 & 1.011 & 0.985 & 0.011 & 1.005 & 1.026 & 1.055 & 1.050 & 0.032 & 1.00 & 1.004 & 1.017 & 1.035& \textbf{0.014} & 1.004 & 1.022 & 1.082 & 1.134 & \underline{0.060}\\
       
       & iTransformer (\citeyear{liu2023itransformer}) & 1.016 & 1.016 & 1.030 & 1.100 & 0.040 & 0.989 & 0.966 & 0.989 & 0.992 & 0.015 & 0.983 & 0.994 & 1.050 & 1.092 & 0.041 & 0.987 & 01.020 & 1.074 & 1.098 & 0.051 & 1.010 & 1.112 & 1.456 & 1.692 & 0.318 \\
       
       & FrNet (\citeyear{zhang2024frnet}) & 0.994 & 0.997 & 1.021 & 1.035 & \underline{0.017} & 1.006 & 1.003 & 1.024 & 1.033 & 0.016 & 1.017 & 1.023 & 1.032 & 1.056 & 0.032 & 1.000 & 1.004 & 1.050 & 1.091 & 0.036 & 1.000 & 1.069 & 1.191 & 1.240 & 0.125\\
       
       \hline
       CNN 
       & ModernTCN (\citeyear{luo2024moderntcn}) & 0.981 & 0.969 & 0.971 & 0.981 & 0.024 & 0.985 & 0.994 & 1.00 & 0.994 & \underline{0.008} & 1.000 & 1.008 & 0.989 & 0.991 & \underline{0.006} & 0.975 & 0.962 & 0.966 & 0.979 & 0.029 & 1.013 & 1.067 & 1.146 & 1.264 & 0.122\\
    \hline
       \multirow{6}{*}{MLP} 
       & DLinear (\citeyear{zeng2023transformers}) & 0.983 & 0.973 & 0.976 & 0.985 & 0.020 & 0.0987 & 0.989 & 1.012 & 1.034 & 0.017 & 1.003 & 1.012 & 1.030 & 1.048 & 0.023 & 1.006 & 1.033 & 1.086 & 1.129 & 0.063 & 0.983 & 1.006 & 1.080 & 1.151 & 0.065\\
       
        & SparseTSF (\citeyear{lin2024sparsetsf}) & 1.000 & 1.019 & 1.058 & 1.067 & 0.036 & 1.014 & 1.023 & 1.032 & 1.055 & 0.031 & 0.997 & 0.997 & 1.000 & 1.011 & \textbf{0.004} & 0.995 & 1.000 & 1.022 & 1.035 & \underline{0.015} & 1.015 & 1.077 & 1.172 & 1.249 & 0.128\\
        
        & FreTS (\citeyear{yi2024frequency}) & 0.998 & 1.007 & 1.021 & 1.052 & 0.020 & 0.931 & 0.944 & 0.898 & 0.991 & 0.058 & 1.024 & 1.013 & 0.993 & 1.006 & 0.012 & 0.956 & 0.930 & 0.890 & 1.072 & 0.073 & 0.992 & 1.103 & 1.110 & 1.148 & 0.092\\
        
        & HADL (w/) & 1.000 & 1.005 & 1.025 & 1.042 & 0.018 & 1.000 & 1.003 & 1.015 & 1.024 & 0.010 & 1.002 & 1.014 & 1.041 & 1.05 & 0.029 & 1.000 & 1.009 & 1.036 & 1.059 & 0.026 & 1.002 & 1.020 & 1.083 & 1.140 & 0.061\\
        
        & HADL(w/o) &  1.000 & 1.002 & 1.017 & 1.035 & \textbf{0.013} & 0.997 & 1.000 & 1.011 & 1.020& \textbf{0.008} & 1.002 & 1.011 & 1.032 &  0.1050 & 0.024 & 1.000 & 1.009 & 1.032 & 1.050 & 0.022 & 1.002 & 1.016 & 1.071 & 1.124 & \textbf{0.053}\\

    \end{tabular} \label{tab: Robustness Testing}
    \end{adjustbox}
\end{table*}

\textbf{Robustness Testing} Table \ref{tab: Robustness Testing} presents robustness testing results for HADL, across different noise intensities compared to baseline models. The full results are in Appendix \ref{appendix sub: Robustness Testing}.

HADL achieves lowest \ac{MAV} in ETTh1 ($0.013$), ETTh2 ($0.008$), and Traffic ($0.053$), indicating a strong resistance to noise. However, it underperforms on ETTm2 ($0.022$), where PatchTST ($0.014$) and SparseTSF ($0.015$) perform better. 

The \ac{NRR} score for HADL remains stable up to $\eta<1.3$ ($1.000$ to $1.016$), while other \ac{MLP} models show greater variability starting from $\eta=0.3$ ($0.983$ to $1.023$) in ETT datasets. Compared to transformers, FrNet shows comparable \ac{NRR} for $\eta<1.3$ ($0.994$ to $1.023$) in ETT datasets, and it degrades significantly in the Traffic dataset ($1.000$ to $1.069$). PatchTST exhibits better noise resilience than other transformers across all datasets, but its \ac{NRR} starts to fluctuate from $\eta\ge0.7$ ($\ge1.007$), whereas HADL remains more stable ($\ge1.000$).

\textbf{Combined Analysis} HADL excels as a lightweight, noise-resilient forecasting model with strong overall accuracy. In ETTm2, it performs well in forecasting while remaining lightweight, but does not perform well in robustness, where PatchTST excels with a larger parameter count ($>700$K). In contrast, on Traffic, HADL struggles in forecasting but achieves the highest robustness.

In particular, in ETTh1 and ETTh2, HADL ranks best in both multivariate forecasting and robustness, showcasing the effectiveness of its design. With strong \ac{MSE} performance at minimal parameter cost and significantly lower FLOPs, HADL emerges as a highly efficient alternative to heavier models like PatchTST, iTransformer, and FrNet while also proving to be a stronger contender against ultra-lightweight models such as SparseTSF.





\section{Ablation Studies and Analysis}

We conduct a series of experiments on ETTh1 and ETTh2 datasets to validate the effects of various components of the HADL framework. We use the rank $=40$ for the low rank layer as part of the test, unless otherwise specified.

\subsection{Effects of Haar Decomposition} \label{sub sec: abalation haar decomposition}
We evaluate the significance of Haar decomposition by comparing the performance of our model with (w/) and without (w/o) this operation, as shown in Table \ref{tab: Abalation Haar Testing}. 

The results indicate that models using Haar decomposition achieve similar or better accuracy compared to the without Haar setup while benefiting from reduced computational complexity and a $20$--$40\%$ reduction in parameter size. This shows that Haar decomposition effectively compresses the input sequence by discarding noisy components, thereby maintaining or enhancing predictive performance.

\begin{table}[h]
    \centering
    \caption{Comparison of \ac{MSE} and count of total trainable parameters for ETTh1 and ETTh2 datasets for lookback window $L=512$ and prediction length $H=\{96, 192, 336, 720\}$ for HADL framework with (w/) and without (w/o) Haar decomposition.}
    \begin{adjustbox}{width=0.4\columnwidth}
    \begin{tabular}{cc||c|c||c|c}
        &\multirow{2}{*}{H} & \multicolumn{2}{c||}{MSE} & \multicolumn{2}{c}{Parameters} \\
        \cline{3-6}
         & & w/ Haar & w/o Haar & w/ Haar & w/o Haar  \\
        \hline
        \multirow{4}{*}{\rotatebox[origin=c]{90}{ETTh1}} & 96 &  0.362 & 0.362 & 14.18K & 24.42K \\ 
        & 192 & 0.402 & 0.397 & 18.11K & 26.35K \\
        & 336 & 0.421 & 0.421 & 24.02K & 34.26K \\
        & 720 & 0.440 & 0.440 & 39.76K & 50.0K \\
         \hline
        \multirow{4}{*}{\rotatebox[origin=c]{90}{ETTh2}} & 96 &  0.271 & 0.276 & 14.18K & 24.42K \\ 
        & 192 & 0.333 & 0.335 & 18.11K & 26.35K \\
        & 336 & 0.359 & 0.366 & 24.02K & 34.26K \\
        & 720 & 0.394 & 0.396 & 39.76K & 50.0K \\
         \hline
    \end{tabular}
    \end{adjustbox}
    \label{tab: Abalation Haar Testing}
\end{table}

We also examine the limitations of Haar decomposition with respect to the length of the lookback window (Appendix \ref{appendix sub: Abalation Haar Testing}). A short lookback window, after Haar decomposition, loses information due to halving the original length, which reduces the model's accuracy. Thus, a sufficiently large or optimal lookback window is necessary for effective performance.

\subsection{Effects of Low-Rank Layer} \label{sub sec: abalation low rank layer}

In this experiment, we evaluate the impact of replacing the low-rank layer with a standard linear layer. As shown in Table \ref{tab: Low Rank Testing}, the low-rank layer consistently outperforms the standard linear layer, achieving lower \ac{MSE} while significantly reducing the parameter count, leading to enhanced computational efficiency. The improvement in \ac{MSE} stems from the low-rank layer's ability to focus on a smaller, more relevant subset of weights, thus minimizing the influence of noisy components in the time series. 

\begin{table}[h]
    \centering
    \caption{Comparison of MSE and count of parameters for ETTh1 and ETTh2 datasets for input sequence length $L=512$ and prediction length $H=\{96, 192, 336, 720\}$ for HADL framework model with Low-Rank($r=40$) and Standard Linear Layer. (\textit{Bias is set to False}.)}
    \begin{adjustbox}{width=0.5\columnwidth}
    \begin{tabular}{cc||c|c||c|c}
        &\multirow{2}{*}{H} & \multicolumn{2}{c||}{MSE} & \multicolumn{2}{c}{Parameters} \\
        \cline{3-6}
         & & Low-Rank & Standard & Low-Rank & Standard  \\
        \hline
        \multirow{4}{*}{\rotatebox[origin=c]{90}{ETTh1}} & 96 &  0.362 & 0.442 & 14.08K & 24.58K \\ 
        & 192 & 0.397 & 0.463 & 17.92K & 49.15K \\
        & 336 & 0.423 & 0.472 & 23.68K & 86.02K \\
        & 720 & 0.420 & 0.487 & 39.04K & 184.32K \\
         \hline
        \multirow{4}{*}{\rotatebox[origin=c]{90}{ETTh2}} & 96 &  0.274 & 0.312 & 14.08K & 24.58K \\
        & 192 & 0.332 & 0.347 & 17.92K & 49.15K \\
        & 336 & 0.364 & 0.361 & 24.02K & 86.35K \\
        & 720 & 0.382 & 0.397 & 39.04K & 184.32K \\
         \hline
    \end{tabular}
    \end{adjustbox}
    \label{tab: Low Rank Testing}
\end{table}

We also examine the weights of the low-rank versus standard linear matrices and observe that the standard linear layer has weights across a broad range of indices, while the low-rank layer has more focused and targeted weights (Appendix \ref{appendix sub: abalation Low Rank Testing}).

The effectiveness of the low-rank layer is highly dependent on the choice of rank. The studies show that the optimal rank is crucial for achieving a balance between model simplicity and predictive performance (Appendix \ref{appendix sub: abalation Low Rank Testing}).

\subsection{Effects of Discrete Cosine Transform} \label{sub sec: abalation discrete cosine transform}

\begin{figure}[h]
    \centering
    \begin{subfigure}[with (w) DCT]{
        \includegraphics[width=0.30\columnwidth, trim={14mm 16mm 4mm 15mm}, clip]{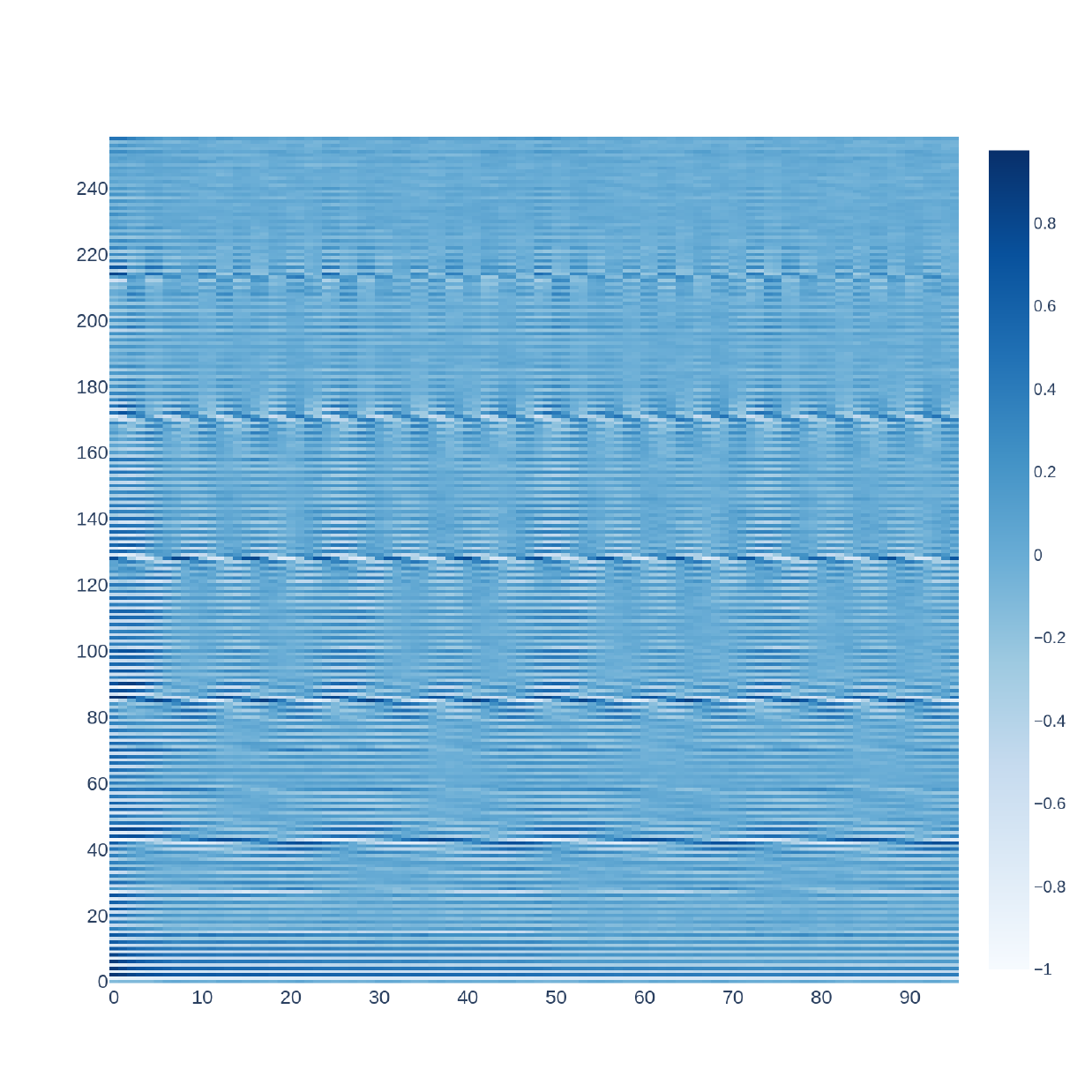}
        }
    \end{subfigure}%
    \begin{subfigure}[without (w/o) DCT]{
        \includegraphics[width=0.30\columnwidth, trim={14mm 16mm 4mm 15mm}, clip]{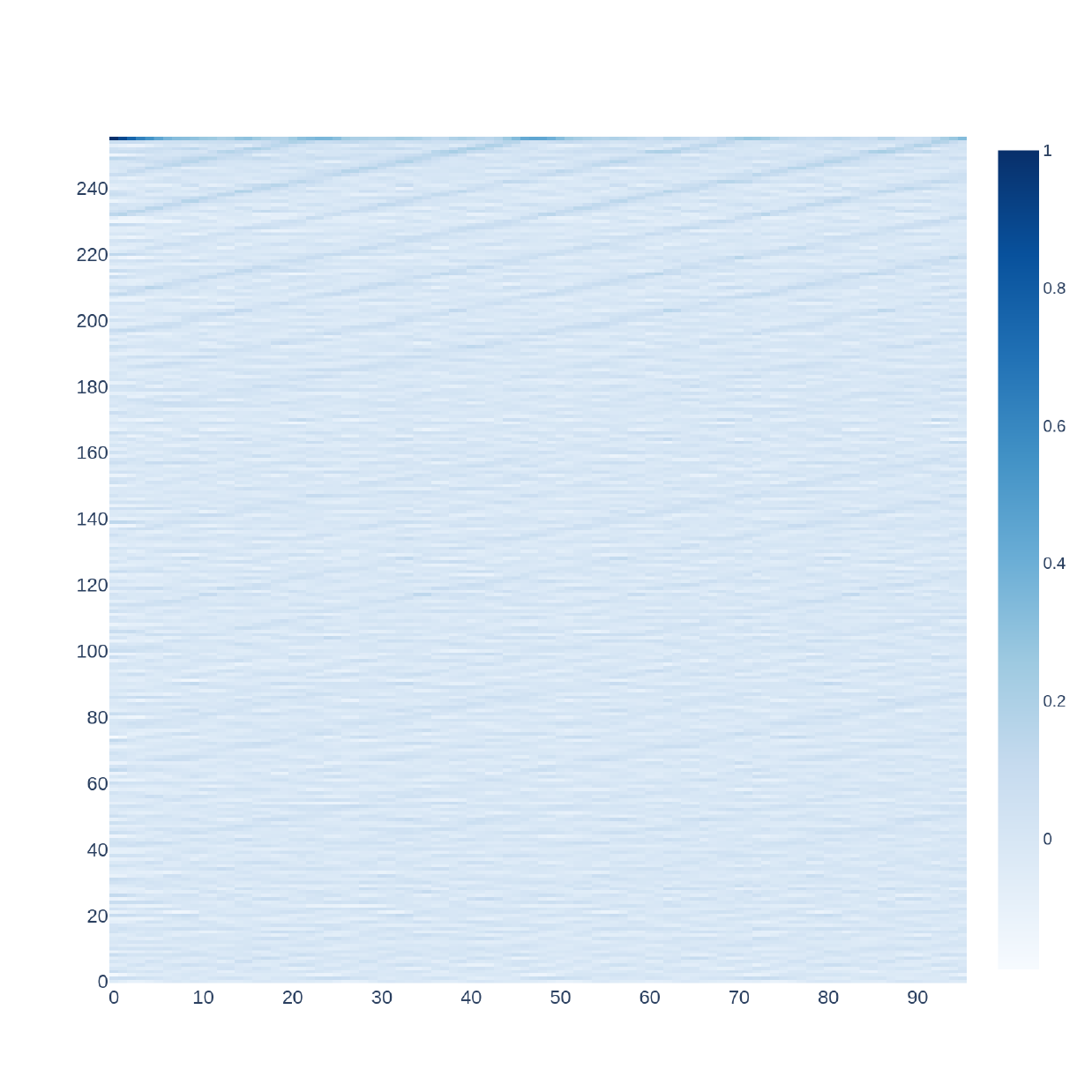}
        }
    \end{subfigure}
    \caption{Heatmap plot of the Low-Rank Matrix for a lookback window $L=512$ and a prediction length $H=96$ in the HADL framework with (w/) and without (w/o) Discrete Cosine Transformation on the ETTh1 dataset.}
    \label{fig: appendix heatmap for DCT matrices}
\end{figure}

In this experiment, we evaluate the impact of bypassing inversion from the frequency domain to the time domain during prediction, as discussed in Section \ref{sub sec: key features}. Figure \ref{fig: appendix heatmap for DCT matrices} presents heat maps of low-rank matrix plots for $L=512$ and $L=96$ in the HADL framework with (w/) and without (w/o) \ac{DCT}, using the ETTh1 dataset. The resulting \ac{MSE} values are $0.362$ (w/ DCT) and $0.371$ (w/o DCT), indicating a slight improvement with \ac{DCT}.

The plots reveal that incorporating \ac{DCT} enhances pattern clarity, distinguishing key structures more effectively than the model without \ac{DCT}. While the \ac{DCT}-enabled model tends to assign more positive weights, the non-\ac{DCT} version exhibits more negative weights. Therefore, both approaches successfully identify and emphasize the critical components of the time series. This suggests that the low-rank layer can effectively learn essential features without requiring explicit inversion from the frequency domain to the time domain, as it inherently captures the extra weights typically introduced during inversion.

Further \ac{MSE} evaluations across various prediction lengths show consistent improvements or comparable performance when using \ac{DCT}, reinforcing its effectiveness in the HADL framework (Appendix \ref{appendix sub: DCT Testing}).

\section{Discussion and Conclusion}
In this work, we introduced HADL, a novel forecasting framework combining \ac{DWT} and \ac{DCT} for noise reduction and robust feature extraction, alongside a lightweight low-rank linear prediction layer, resulting in an efficient, noise-resilient, and accurate model for long-term multivariate time series forecasting.

However, there are several limitations that present opportunities for future improvement. One significant limitation of the HADL framework is its sensitivity to the size of the lookback window. While the model performs well with medium to large lookback windows, its accuracy degrades when the window is too small, primarily due to the reduced amount of historical information available for prediction. Conversely, when the lookback window is extended to ultra-long ranges, the model could potentially benefit from incorporating multiple levels of Haar decomposition. However, this introduces challenges, such as an increase in the number of trainable parameters. Additionally, the current strategy of discarding detail coefficients at each decomposition level must be revisited, as this may lead to the loss of important fine-grained information.

Despite these limitations, HADL outperforms ultra-lightweight models and remains highly efficient compared to transformer-based and \ac{MLP} models, all while maintaining impressive robustness to noise and providing stable performance across varying noise levels. In conclusion, HADL framework offers a strong balance between forecasting accuracy, parameter efficiency, and noise resilience, making it a highly competitive solution for long-term multivariate time series forecasting.

\bibliographystyle{unsrtnat}
\bibliography{main}  

\newpage
\appendix
\onecolumn

\section*{\Huge Supplementary Material}
\vspace{1em}

\section{Benchmark Datasets} \label{appendix: Datasets}

\begin{table}[h]
    \centering
    \caption{Description of Benchmark Datasets}
    \begin{tabular}{c|ccccccc}
        Dataset & ETTh1 & ETTh2 & ETTm1 & ETTm2 & Weather & Traffic & Electricity \\
        \hline
        Features & 7 & 7 & 7 & 7 & 21 & 862 & 321 \\
        Timesteps & 17420 & 17420 & 69680 & 69680 & 52697 & 17544 & 26304 \\
        Granularity & 1 hour & 1 hour & 15 min & 15 min & 10 min & 1 hour & 15 min \\
        \hline
    \end{tabular}
    \label{tab: appendix dataset information}
\end{table}

Below is a description of benchmark datasets used: 
 \begin{itemize}
     \item ETT\footnote{https://github.com/zhouhaoyi/ETDataset} \cite{zhou2021informer}: The Electricity Transformer Temperature captures station's load and oil temperature recorded every 1 hour(ETTh1, ETTh2) and 15 mins(ETTm1, ETTm2) for two separate counties in China over a period of 2 years.
     \item Electricity\footnote{https://github.com/laiguokun/multivariate-time-series-data} \cite{lai2018modeling}: The Electricity Consumption Load is a collection of power consumption recorded in kWh for each household with 15 mins sampling rate. There are 321 household consumption recorded.
    \item Weather\footnote{https://www.bgc-jena.mpg.de/wetter/} \cite{zhou2021informer}: The climatological data consists of 21 meteorological features such as air temperature, humidity, etc. captured every 10 minutes for the year 2020.  
    \item Traffic\footnote{https://github.com/laiguokun/multivariate-time-series-data} \cite{lai2018modeling}: The data from California Department of Transportation describes road occupancy rates(between 0 and 1) measured by 862 different sensors on San Franscisco Bay freeways hourly over a period of 48 months.
 \end{itemize}

\clearpage
\section{Alogrithm} \label{appendix: algorithm}

The algorithm formally summarizes the necessary steps for the HADL framework.

\begin{algorithm}[h]
\caption{HADL Time Series Forecasting}
\label{alg:haar-dct}
\begin{algorithmic}[1]
\STATE \textbf{Input:} Multivariate time series $X_T \in \mathbb{R}^{C \times L}$, lookback window $L$, horizon length $H$
\STATE \textbf{Output:} Predicted values $\hat{Y}_T \in \mathbb{R}^{C \times H}$

\STATE \textbf{Step 1: Wavelet Decomposition}
\STATE Apply Discrete Wavelet Transform (DWT) with Haar Wavelet:
\[
A_T, \_ \gets \text{DWT}(X_T)
\]
\STATE Retain approximation coefficients $A_T \in \mathbb{R}^{C \times \frac{L}{2}}$ to reduce noise.

\STATE \textbf{Step 2: Frequency Domain Transformation}
\STATE Transform approximation coefficients into the frequency domain using Discrete Cosine Transform (DCT):
\[
A_F \gets \text{DCT}(A_T)
\]
\STATE Normalize the frequency representation:
\[
A_F \gets \frac{2}{L} \cdot A_F
\]

\STATE \textbf{Step 3: Low-Rank Prediction}
\STATE Map the frequency representation $A_F \in \mathbb{R}^{C \times \frac{L}{2}}$ to the target forecast in the time domain using a low-rank layer:
\[
\hat{Y}_T \gets \text{LowRankLayer}(A_F)
\]
\STATE \textbf{Return:} Predicted values $\hat{Y}_T$
\end{algorithmic}
\end{algorithm}

\clearpage
\section{Additional Ablation Studies} \label{appendix: Abalation studies}

\subsection{Haar decomposition Testing} \label{appendix sub: Abalation Haar Testing}
Table \ref{tab: appendix Abalation Haar sequence length} presents the results of the HADL framework for lookback windows $L=\{48,96,192, 336,512,720\}$ and prediction lengths $H=\{96,192, 336,720\}$ on the ETTh1 and ETTh2 datasets with  rank $r=40$. The results highlight a limitation of the HADL for smaller lookback windows, where accuracy deteriorates. This is likely due to the halving of the original time series length, which may result in the loss of critical information necessary for accurate predictions when the lookback window is too small.

\begin{table}[h]
    \centering
    \caption{Results of the HADL framework in \ac{MSE} for lookback windows $L=\{48,96,192, 336,512,720\}$ and prediction lengths $H=\{96,192, 336,720\}$}
    \begin{adjustbox}{width=0.4\columnwidth}
    \begin{tabular}{c|c|c|c|c|c}
    
         & \diagbox[width=2cm]{L}{H}  & 96 & 192 & 336 & 720 \\
         \hline
         \multirow{6}{*}{\rotatebox[origin=c]{90}{ETTh1}}& 48 & 0.400 & 0.452 & 0.502 & 0.501 \\
         & 96 & 0.385 & 0.435 & 0.476 & 0.470 \\
         & 192 & 0.377 & 0.422 & 0.451 & 0.443 \\
         & 336 & 0.368 & 0.403 & 0.427 & 0.432 \\
         & 512 & 0.362 & 0.402 & 0.421 & 0.440 \\
         & 720 & 0.371 & 0.405 & 0.442 & 0.454 \\
         \hline
         \multirow{6}{*}{\rotatebox[origin=c]{90}{ETTh2}} & 48 & 0.302 & 0.394 & 0.445 & 0.450 \\
         & 96 & 0.290 & 0.375 & 0.414 & 0.421 \\
         & 192 & 0.284 & 0.356 & 0.385 & 0.409 \\
         & 336 & 0.275 & 0.336 & 0.361 & 0.397 \\
         & 512 & 0.271 & 0.333 & 0.359 & 0.394 \\
         & 720 & 0.274 & 0.335 & 0.360 & 0.398 \\
         \hline
    \end{tabular}
    \end{adjustbox}
    \label{tab: appendix Abalation Haar sequence length}
\end{table}

Table \ref{tab: appendix Varying Low Rank Testing} presents the results for varying ranks $r= \{15, 35, 55, 75\}$ in the low-rank layer, with a window size of  $L=512$ and prediction lengths $H=\{96, 192, 336, 720\}$ for the ETTh1 and ETTh2 datasets. The results suggest that a very low rank reduces accuracy due to a reduced number of parameters. An optimal rank strikes a balance, providing good accuracy without excessive memory requirements.

\begin{table*}[h]
    \centering
    \caption{Results of the HADL framework in \ac{MSE} for a lookback window of $L=512$, prediction lengths $H=\{96, 192, 336, 720\}$ and Low Rank Layer with rank $r= \{15, 35, 55, 75\}$ for ETTh1 and ETTh2 datasets.}
    \begin{adjustbox}{width=0.8\textwidth}
    \begin{tabular}{c|c||c|c|c|c||c|c|c|c}
        & & \multicolumn{4}{c||}{MSE} & \multicolumn{4}{c}{Parameters} \\
        \cline{3-10}
        & \diagbox[width=2cm, height=0.5cm]{H}{r} & 15 & 35 & 55 & 75 & 15 & 35 & 55 & 75 \\
        \hline
        \multirow{4}{*}{\rotatebox[origin=c]{90}{ETTh1}} & 96 & 0.367 & 0.369 & 0.362 & 0.363 & 5.38K & 12.42K & 19.46K & 26.5K \\
        & 192 & 0.401 & 0.404 & 0.395 & 0.395 & 6.91K & 15.87K & 24.83K & 33.79K \\
        & 336 & 0.426 & 0.422 & 0.421 & 0.421 & 9.22K & 21.06K & 32.9K & 44.74K \\
        & 720 & 0.445 & 0.445 & 0.442 & 0.435 & 15.36K & 34.88K & 54.4K & 73.92K \\
         \hline    
         \multirow{4}{*}{\rotatebox[origin=c]{90}{ETTh2}} & 96 & 0.274 & 0.270 & 0.270 & 0.271 & 5.38K & 12.42K & 19.46K & 26.5K \\
        & 192 & 0.335 & 0.334 & 0.333 & 0.331 & 6.91K & 15.87K & 24.83K & 33.79K \\
        & 336 & 0.364 & 0.359 & 0.358 & 0.357 & 9.22K & 21.06K & 32.9K & 44.74K \\
        & 720 & 0.398 & 0.396 & 0.393 & 0.392 & 15.36K & 34.88K & 54.4K & 73.92K \\
         \hline     
    \end{tabular}
    \end{adjustbox}
    \label{tab: appendix Varying Low Rank Testing}
\end{table*}

\subsection{Low Rank Testing} \label{appendix sub: abalation Low Rank Testing}

\begin{figure}[h]
    \centering
    \begin{subfigure}[Low Rank]{
        \includegraphics[width=0.3\columnwidth, trim={14mm 16mm 4mm 15mm}, clip]{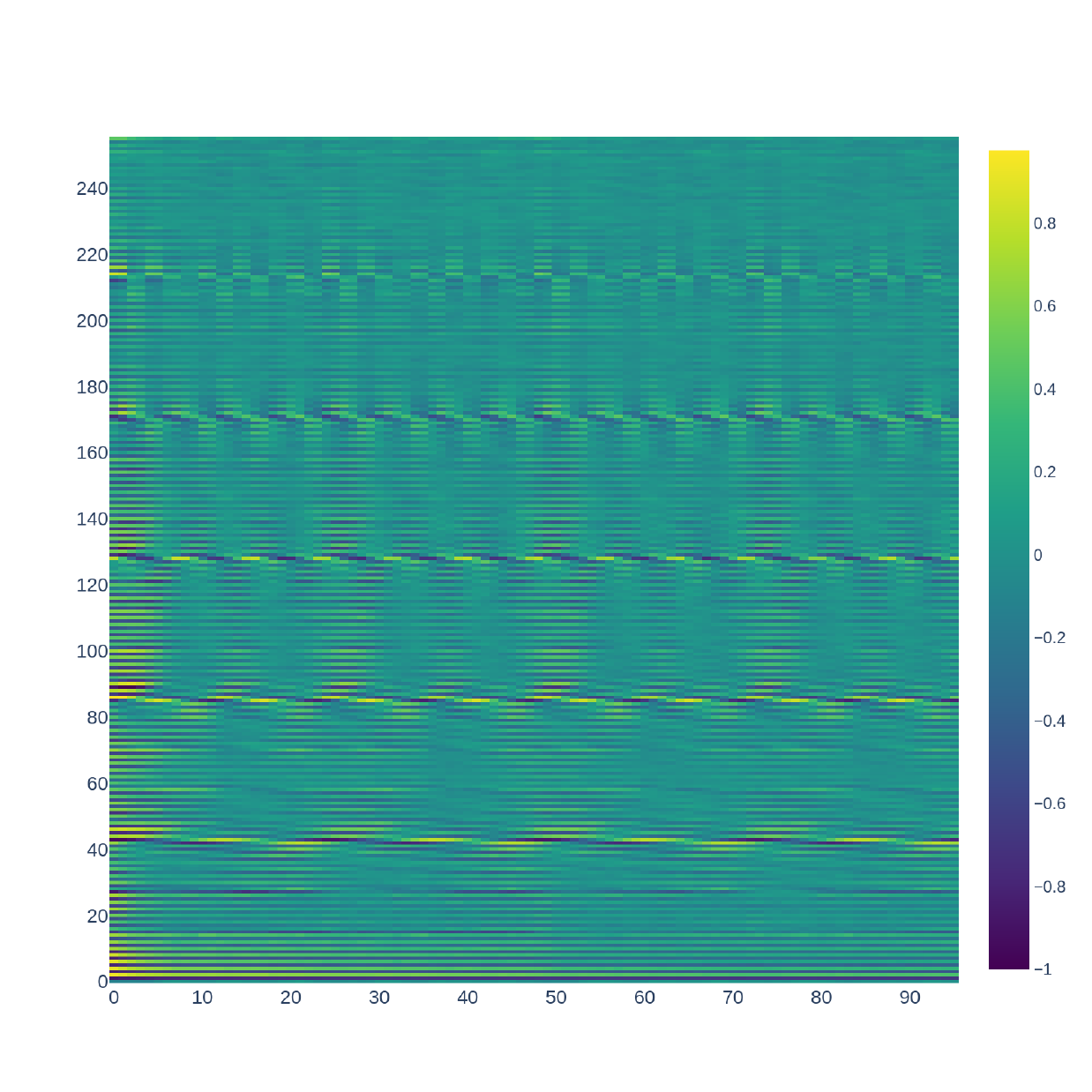}
        }
    \end{subfigure}%
    \begin{subfigure}[Standard]{
        \includegraphics[width=0.3\columnwidth, trim={14mm 16mm 4mm 15mm}, clip]{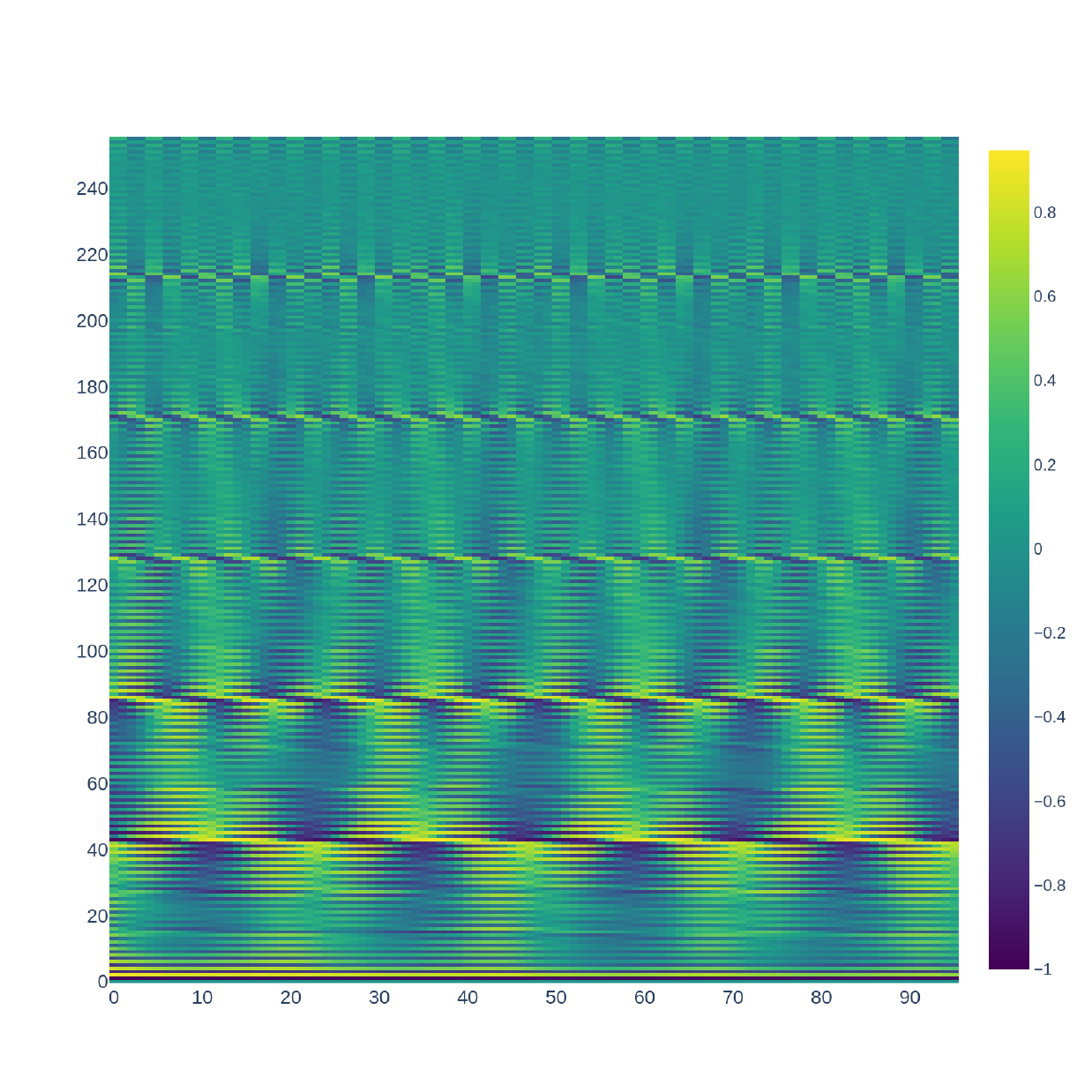}
        }
    \end{subfigure}
    
    \caption{Heatmap plot for HADL framework comparing the Standard Linear and Low-Rank Matrices for a lookback window $L=512$ and a prediction length $H=96$ for ETTh1 dataset. Bias is set to \textit{False}.}
    \label{fig: appendix heatmap plot of low rank and standard}
\end{figure}

Figure \ref{fig: appendix heatmap plot of low rank and standard} presents a heatmap comparing low-rank and standard linear matrices for $L=512$ and $H=96$ in the HADL framework, without bias, for the ETTh1 dataset. The results indicate that the low-rank approach assigns weights less aggressively compared to the standard linear layer, thereby reducing the influence of noisy components that are weighted more heavily in the standard linear layer.

Figure \ref{fig: appendix heatmap plot of low rank and standard} and Table \ref{tab: Low Rank Testing} demonstrate that the generalization capability of the low-rank approach helps mitigate noise in time series data. Moreover, it achieves comparable outputs with fewer parameters and lower memory requirements.

\subsection{Discrete Cosine Transform Testing} \label{appendix sub: DCT Testing}
We investigate the impact of incorporating the \ac{DCT} in our model by removing the \ac{DCT} operation and evaluating the model's \ac{MSE} based on predictions directly from the time domain. Without the \ac{DCT}, the low-rank layer relies solely on the input transformed by the Haar decomposition, which operates in the time domain instead of the frequency domain.

\begin{table}[H]
    \centering
    \caption{\ac{MSE} results for the HADL framework with (w/) and without (w/o) Discrete Cosine Transformation, for a lookback window $L=512$ and prediction lengths $H=\{96, 192, 336, 720\}$ on the ETTh1 and ETTh2 datasets.}
    \begin{adjustbox}{width=0.30\textwidth}
    \begin{tabular}{cc||c|c}
         & \multirow{2}{*}{H} & \multicolumn{2}{c}{MSE}  \\
         \cline{3-4}
         & & w/ DCT & w/o DCT \\
        \hline
         \multirow{4}{*}{\rotatebox[origin=c]{90}{ETTh1}} & 96 & 0.362 & 0.371 \\
         & 192 & 0.402 & 0.402 \\
         & 336 & 0.421 & 0.432 \\
         & 720 & 0.440 & 0.438 \\
         \hline 
         \multirow{4}{*}{\rotatebox[origin=c]{90}{ETTh2}} & 96 & 0.271 & 0.277 \\
         & 192 & 0.333 & 0.337 \\
         & 336 & 0.359 & 0.359 \\
         & 720 & 0.394 & 0.396 \\
         \hline
    \end{tabular}
    \end{adjustbox} 
    \label{tab: DCT Testing}
\end{table}

The results in Table \ref{tab: DCT Testing} indicate that the model with \ac{DCT} either outperforms or achieves similar \ac{MSE} values compared to the model without \ac{DCT} across both datasets.

\clearpage
\section{Results} \label{appendix: Results}

\subsection{Multivariate Forecasting} \label{appendix sub: Multivariate Forecasting results}

Table \ref{tab: Appendix Multivariate Forecasting} presents the results of multivariate long-term time series forecasting, evaluated using \ac{MSE} and \ac{MAE}, across all benchmark datasets. It compares the performance of \ac{SOTA} models with a lookback window of $L=512$ and prediction lengths $H = \{96, 192, 336, 720\}$ to that of the HADL framework, both with (w/) and without (w/o) L1 regularization. All models were trained for 100 epochs with a patience level of 20.

\begin{table*}[htbp]
    \centering 
    \caption{Results in \ac{MSE} and \ac{MAE} (lower is better) for multivariate long-term time series forecasting on benchmark datasets. We compare against baseline models using a standard prediction length $H = \{96, 192, 336, 720\}$ and a lookback window $L=512$ for all models.}
    \begin{adjustbox}{width=\textwidth}
    \renewcommand{\arraystretch}{1.5}
    \begin{tabular}{cc|cc|cc|cc|cc|cc|cc|cc|cc|cc}
    \multicolumn{2}{c|}{} & \multicolumn{10}{c|}{MLP} & \multicolumn{6}{c|}{Transformers} & \multicolumn{2}{c}{CNN} \\
    \cline{3-20}
     & & \multicolumn{2}{c|}{HADL (w/)} & \multicolumn{2}{c|}{HADL (w/o)} & \multicolumn{2}{c|}{DLinear (\citeyear{zeng2023transformers})} & \multicolumn{2}{c|}{SparseTSF (\citeyear{lin2024sparsetsf})} & \multicolumn{2}{c|}{FreTS (\citeyear{yi2024frequency})} & \multicolumn{2}{c|}{PatchTST (\citeyear{nie2022time})} & \multicolumn{2}{c|}{iTransformer (\citeyear{liu2023itransformer})} & \multicolumn{2}{c|}{FrNet (\citeyear{zhang2024frnet})} & \multicolumn{2}{c}{ModernTCN (\citeyear{luo2024moderntcn})}\\
     Dataset & H & MSE & MAE & MSE & MAE & MSE & MAE & MSE & MAE & MSE & MAE & MSE & MAE & MSE & MAE & MSE & MAE & MSE & MAE\\
     \hline
     \multirow{4}{*}{ETTh1} & 96 & 0.363 & 0.393 & 0.365 & 0.394 & 0.382 & 0.403 & 0.392 & 0.408 & 0.396 & 0.420 & 0.394 & 0.419 & 0.399 & 0.423 &0.362 &0.392 & 0.385 & 0.411\\
     & 192 & 0.395 & 0.414 & 0.397 & 0.414 & 0.418 & 0.426  & 0.412 & 0.421 & 0.435 & 0.448 & 0.424 & 0.440 & 0.428 & 0.444 &0.390 &0.415 & 0.421 & 0.430\\
     & 336 & 0.421 & 0.429 & 0.423 & 0.429 & 0.455 & 0.469 & 0.425 & 0.432 & 0.471 & 0.472 & 0.457 & 0.467 & 0.472 & 0.474 &0.427 & 0.436 & 0.448 & 0.448\\
     & 720 & 0.444 & 0.462 & 0.446 & 0.462 & 0.496 & 0.498 & 0.425 & 0.453 & 0.565 & 0.551& 0.461 & 0.475 & 0.531 & 0.520 &0.468 &0.478 &0.531 & 0.511\\
     \hline
     \multirow{4}{*}{ETTh2} & 96 & 0.271 & 0.336 & 0.272 & 0.335 & 0.354 & 0.391 & 0.285 & 0.345 & 0.318 & 0.383 & 0.292 & 0.355 & 0.301 & 0.359 & 0.269 & 0.336 &0.300 & 0.356\\
     & 192 & 0.334 & 0.378 & 0.334 & 0.377 & 0.512 & 0.481 & 0.345 & 0.382 & 0.432 & 0.453 & 0.347 & 0.396 & 0.381 & 0.408& 0.330& 0.378&0.348 & 0.389\\
     & 336 & 0.358 & 0.403 & 0.358 & 0.402 & 0.603 & 0.536 & 0.364 & 0.400 & 0.479 & 0.483 & 0.380 & 0.423 &  0.395 & 0.427&0.358 &0.405 & 0.384 & 0.420\\
     & 720 & 0.395 & 0.440 & 0.395 & 0.439 & 1.060 & 0.722 & 0.384 & 0.424 & 0.809 & 0.650 & 0.431 & 0.453 & 0.460 & 0.468&0.391 &0.432 & 0.410 & 0.440\\
     \hline
     \multirow{4}{*}{ETTm1} & 96 & 0.304 & 0.349 & 0.306 & 0.349 & 0.292 & 0.338 & 0.330 & 0.365 & 0.317 & 0.360 & 0.293 & 0.344 & 0.315 & 0.367 & 0.295 &0.347 &0.310 & 0.361\\
     & 192 & 0.337 & 0.367 & 0.339& 0.368 & 0.333 & 0.362 & 0.347 & 0.372 & 0.356 & 0.384 & 0.335 & 0.367 & 0.350 & 0.389 &0.337 & 0.371 & 0.354 & 0.387\\
     & 336 & 0.365 & 0.383 & 0.367 & 0.383 & 0.368 & 0.382 & 0.374 & 0.388 & 0.385 & 0.404 & 0.368 & 0.387 & 394 & 0.416 &0.363 & 0.380 & 0.392 & 0.402\\
     & 720 & 0.418 & 0.414 & 0.420 & 0.413 & 0.428 & 0.417 & 0.424 & 0.415 & 0.440 & 0.434 & 0.423 & 0.420 & 0.446 & 0.445& 0.421 & 0.415 & 0.446 & 0.435\\
     \hline
     \multirow{4}{*}{ETTm2} & 96 & 0.163 & 0.253 & 0.163 & 0.252 &0.204 & 0.288 &0.173 & 0.260 & 0.179 & 0.263 & 0.164 & 0.254 & 0.188 & 0.275 & 0.165 & 0.254 & 0.169 & 0.261\\
     & 192 & 0.218 & 0.290 & 0.218 & 0.290 & 0.333 & 0.362 &0.225 & 0.294 & 0.254 & 0.313 & 0.223 & 0.294 & 0.250 & 0.317& 0.221 & 0.293 & 0.232 & 0.303\\
     & 336 & 0.271 & 0.325 &0.271 & 0.325 & 0.342 & 0.386 & 0.277 & 0.328 & 0.314 & 0.354 & 0.277 & 0.329 & 0.326 & 0.361 & 0.274 & 0.328 & 0.307 & 0.349\\
     & 720 & 0.359 & 0.382 & 0.359& 0.382 & 0.570 & 0.488 & 0.362 & 0.382 & 0.389 & 0.412 & 0.381 & 0.395 & 0.391 & 0.408 & 0.360 & 0.387 & 0.385 & 0.399\\
     \hline
     \multirow{4}{*}{Traffic} & 96 & 0.413 & 0.299 & 0.413 & 0.294 &  0.418 & 0.298 & 0.413 & 0.299 & 0.398 & 0.278 & 0.489 & 0.354 & 0.433 & 0.324 & 0.367 & 0.274 &0.392 & 0.285\\
     & 192 & 0.433 & 0.312 & 0.433 & 0.308 & 0.430 & 0.303 & 0.433 & 0.312 & 0.414 & 0.285 & 0.498 & 0.357 & 0.462 & 0.343& 0.390 & 0.286 &0.406 & 0.291\\
     & 336 & 0.445 & 0.317 & 0.444 & 0.314 & 0.440 & 0.309 & 0.445 & 0.317 & 0.428 &0.291 & 0.506 & 0.360 & 0.443 & 0.323& 0.397 & 0.283 & 0.415 & 0.296\\
     & 720 & 0.481 & 0.335 &  0.481 &0.332 & 0.479 & 0.331 & 0.481 & 0.335 & 0.471 & 0.313 & 0.541 & 0.376 & 0.489 & 0.349& 0.433&0.301 &0.464 & 0.327\\
     \hline
     \multirow{4}{*}{Weather} & 96 & 0.171 & 0.224 & 0.170 & 0.223 & 0.142 & 0.204 & 0.179 & 0.233 & 0.181 & 0.252 & 0.145 & 0.198 & 0.158 & 0.210  & 0.143 & 0.196 & 0.156 & 0.212 \\
     & 192 & 0.215 & 0.261 & 0.214 & 0.260 & 0.186 & 0.251 & 0.220 & 0.265 & 0.222& 0.290& 0.189 & 0.239 & 0.203 & 0.249 &0.185 & 0.236 & 0.202 & 0.253\\
     & 336 & 0.260 & 0.295 & 0.260 & 0.295 &0.236 & 0.294 & 0.264 & 0.298 & 0.278& 0.330 & 0.239 & 0.278 & 0.263 & 0.295 & 0.234 & 0.274 & 0.255 & 0.294\\
     & 720  & 0.326 & 0.343 & 0.326 &0.342 & 0.311 & 0.354 & 0.328 & 0.343 & 0.346& 0.386& 0.311 & 0.331 & 0.329 & 0.344 &0.309 & 0.330 & 0.347 & 0.354\\
     \hline
     \multirow{4}{*}{Electricity} & 96 & 0.151 & 0.256 & 0.149 & 0.252 & 0.132 & 0.229 & 0.151 & 0.248 & 0.191 & 0.310 & 0.157 & 0.265 & 0.148 & 0.250 & 0.132 & 0.227 & 0.139& 0.242\\
     & 192 & 0.168 & 0.271 & 0.166 & 0.267 & 0.148 & 0.245 & 0.164 & 0.259 & 0.215 & 0.331 & 0.172 & 0.277 & 0.168 & 0.270 & 0.147 & 0.239 & 0.154 & 0.253\\
     & 336 & 0.183 & 0.286 & 0.181 & 0.282 & 0.164 & 0.263 & 0.180 & 0.278 & 0.258 & 0.368 & 0.188 & 0.291 & 0.175 & 0.274 & 0.163 & 0.257 & 0.172 & 0.275 \\
     & 720 & 0.222 & 0.317 & 0.220 & 0.313 & 0.201 & 0.297 & 0.218 & 0.308 & 0.276 & 0.378 & 0.224 & 0.320 & 0.216 & 0.309 & 0.200 &  0.290 & 0.209 & 0.301\\
     \hline    
        \label{tab: Appendix Multivariate Forecasting}
    \end{tabular}
    \end{adjustbox}   
\end{table*}

\clearpage
\subsection{Univariate Long-Term Forecasting} \label{appendix sub: univariate forecasting results}

Table \ref{tab: Appendix Univariate Forecasting} presents the results of univariate long-term time series forecasting, evaluated using \ac{MSE} and \ac{MAE}, across all benchmark datasets. The table compares \ac{SOTA} models with a lookback window of $L=512$ and prediction lengths of $H = \{96, 192, 336, 720\}$ to the performance of the HADL framework, both with (w/) and without (w/o) L1 regularization. All models are trained for 50 epochs with a patience level of 10.
\begin{table*}[h]
    \centering
    
    \caption{Results in \ac{MSE} and \ac{MAE} (lower is better) for univariate long-term time series forecasting on benchmark datasets. We compare with baseline models with standard prediction length $H = \{96, 192, 336, 720\}$ and input sequence $L=512$ for all models.}
    \begin{adjustbox}{width=\textwidth}
    \renewcommand{\arraystretch}{1.5}
    \begin{tabular}{cc|cc|cc|cc|cc|cc|cc|cc|cc|cc}
    \multicolumn{2}{c|}{} & \multicolumn{10}{c|}{MLP} & \multicolumn{6}{c|}{Transformers} & \multicolumn{2}{c}{CNN} \\
    \cline{3-20}
     & & \multicolumn{2}{c|}{HADL (w/)} & \multicolumn{2}{c|}{HADL (w/o)} & \multicolumn{2}{c|}{DLinear (\citeyear{zeng2023transformers})} & \multicolumn{2}{c|}{SparseTSF (\citeyear{lin2024sparsetsf})} & \multicolumn{2}{c|}{FreTS (\citeyear{yi2024frequency})} & \multicolumn{2}{c|}{PatchTST (\citeyear{nie2022time})} & \multicolumn{2}{c|}{iTransformer (\citeyear{liu2023itransformer})} & \multicolumn{2}{c|}{FrNet (\citeyear{zhang2024frnet})} & \multicolumn{2}{c}{ModernTCN (\citeyear{luo2024moderntcn})}\\
     Dataset & H & MSE & MAE & MSE & MAE & MSE & MAE & MSE & MAE & MSE & MAE & MSE & MAE & MSE & MAE & MSE & MAE & MSE & MAE\\
     \hline
     \multirow{4}{*}{ETTh1} & 96 & 0.101 & 0.244 & 0.100 & 0.244 & 0.064 & 0.191 & 0.064 & 0.199& 0.095 & 0.241 & 0.069 & 0.205 & 0.085 & 0.230 &0.054 & 0.180 & 0.080 & 0.223\\
     & 192 & 0.116 & 0.264 & 0.116 & 0.264 & 0.085 & 0.223 & 0.080 & 0.223& 0.114 & 0.266 & 0.083 & 0.227 & 0.091 & 0.239 &0.067 & 0.204 & 0.085 & 0.236\\
     & 336 & 0.129 & 0.282 & 0.127 &0.280 & 0.108 & 0.257 & 0.090 & 0.240 & 0.122 & 0.277 &0.091 & 0.242 & 0.099 & 0.249 & 0.080 & 0.225 & 0.088 & 0.236\\
     & 720 & 0.167 & 0.322 & 0.169 & 0.324 & 0.191 & 0.361 & 0.107 & 0.256 & 1.794 & 0.333 & 0.106 & 0.256 & 0.131 & 0.289 & 0.095 & 0.240 & 0.108 & 0.262\\
     \hline
     \multirow{4}{*}{ETTh2} & 96 & 0.189 & 0.346 & 0.202 & 0.361 & 0.139 & 0.290  & 0.152 & 0.309 & 0.179 & 0.333 &0.160 & 0.318 & 0.278 & 0.430 &0.132 & 0.283 & 0.245 & 0.404\\
     & 192 &  0.225 & 0.383 & 0.235 & 0.392 & 0.181 & 0.336 & 0.187 & 0.347 & 0.227 & 0.384 & 0.192 & 0.352 & 0.296 & 0.445 & 0.168 & 0.324 & 0.258 & 0.416\\
     & 336 & 0.255 & 0.411 & 0.263 & 0.420 & 0.204 & 0.364 & 0.204 & 0.366 & 0.316 & 0.440 & 0.209 & 0.372 & 0.312 & 0.456 & 0.178 & 0.343 & 0.270 & 0.426\\
     & 720 & 0.339 & 0.474 & 0.364 & 0.493 & 0.334 & 0.470  & 0.241 & 0.394 & 0.325 & 0.464 & 0.252 & 0.406 & 0.405 & 0.519 & 0.221 & 0.376 & 0.347 & 0.481\\
     \hline
     \multirow{4}{*}{ETTm1} & 96 & 0.026 & 0.123 & 0.026& 0.122 &0.026 & 0.122 & 0.028 & 0.130 &1.679 & 0.703 & 0.027 & 0.126 & 0.051 & 0.176 & 0.026 & 0.122 &0.0384 & 0152\\
     & 192 & 0.039 & 0.150 & 0.039 & 0.150 &0.041 & 0.151 & 0.041 & 0.155 & 4.594 & 1.156 & 0.040&  0.153& 0.061 & 0.191 &0.040 & 0.152 & 0.049 & 0.171\\
     & 336 & 0.052 & 0.173 & 0.051& 0.173 & 0.055 & 0.175 & 0.053 & 0.176 & 2.945 & 0.879 & 0.053 & 0.175 & 0.067 & 0.199 & 0.053 & 0.175 & 0.060 & 0.188\\
     & 720 & 0.070 & 0.203 & 0.070 & 0.204 & 0.075 & 0.205 &0.072 & 0.206 & 8.60 & 1.518 & 0.072 & 0.205 & 0.085 & 0.226& 0.072 & 0.206 & 0.077 & 0.215\\
     \hline
     \multirow{4}{*}{ETTm2} & 96 & 0.063 & 0.185 & 0.063 & 0.185 &0.063 & 0.183 &0.071 & 0.202 & 0.140 & 0.264 & 0.069 & 0.199 &0.077 & 0.213 & 0.065 & 0.188 &  0.097 & 0.243 \\
     & 192 & 0.090 & 0.226 & 0.90&0.226 &0.091 & 0.226&0.096& 0.238 & 0.565&0.418 & 0.095 & 0.235 & 0.105 &0.253 & 0.094 & 0.232 & 0.118 & 0.267\\
     & 336 & 0.118 & 0.261 &0.118 & 0.261 & 0.120 & 0.263&0.122 & 0.269 & 1.615& 0.601 & 0.121& 0.267 &0.134 & 0.287 & 0.124 & 0.271 & 0.140 & 0.292\\
     & 720 & 0.171 & 0.321 & 0.171 & 0.321 & 0.174 & 0.321&0.175 & 0.325 & 5.742 & 1.048 & 0.173 & 0.323 &0.186 &0.344 & 0.178 & 0.329 & 0.190 & 0.342\\
     \hline
     \multirow{4}{*}{Electricity} & 96 & 0.227 & 0.342 & 0.245& 0.358& 0.192 & 0.303 &0.234 & 0.343 & 0.287 & 0.398 &0.258 & 0.384 & 0.237 & 0.345 & 0.210 & 0.314 & 0.196 & 0.307\\
     & 192 & 0.263 & 0.367 & 0.283&0.385 & 0.225 & 0.326 &0.262 & 0.362 & 0.331&0.425 &0.287 & 0.384& 0.276 & 0374 & 0.244 & 0.341 & 0.231 & 0.331\\
     & 336 & 0.297 & 0.393 & 0.314 & 0.408 & 0.260 & 0.355&0.289 & 0.383 &0.393 &0.468 &0.314 & 0.404 & 0.283 & 0.382 & 0.285 & 0.372 & 0.270 & 0.361\\
     & 720 & 0.344 & 0.440 &0.364 & 0.456 & 0.294 & 0.397&0.339 &0.433 &0.722 &0.549 &0.358 & 0.449&0.331 & 0.427 & 0.351 & 0.442 & 0.317 & 0.411\\
     \hline
     \multirow{4}{*}{Weather} & 96 & 0.001 & 0.030 & 0.001 & 0.024 &0.005 &0.058 & 0.001 & 0.024 &0.005 & 0.061 &0.001 & 0.027 &0.001 & 0.275 & 0.001 & 0.025 & 0.001 & 0.025\\
     & 192 & 0.001 & 0.031 & 0.001 & 0.0026 &0.005 &0.062 & 0.001 & 0.027& 0.006& 0.064 &0.001 & 0.028 & 0.001 & 0.029 & 0.001 & 0.027& 0.001 & 0.028\\
     & 336 & 0.001 & 0.032 &  0.001 &0.028 & 0.006  & 0.065 &  0.001 & 0.028 & 0.006 & 0.065 & 0.001 & 0.030 & 0.001 & 0.030 & 0.001 & 0.029 & 0.001 & 0.030\\
     & 720 & 0.002 & 0.036 & 0.002 & 0.033 &0.006 & 0.069 & 0.002 & 0.033 & 0.006& 0.068 & 0.002&  0.035& 0.002 & 0.035 & 0.002 & 0.034 & 0.002 & 0.035\\
     \hline
     \multirow{4}{*}{Traffic} & 96 & 0.308 & 0.417 & 0.304 & 0.408 &0.118 & 0.197&0.276  & 0.377 & 3.184& 1.017 & 0.297 & 0.396 & 0.251 & 0.356 & 0.138 & 0.231 & 0.707 & 0.649\\
     & 192 & 0.315 & 0.422 & 0.311 & 0.412 &0.120 & 0.199 & 0.275 & 0.375 & 3.27 & 1.13 &0.297 & 0.396 & 0.275 & 0.375 & 0.134 & 0.223 & 0.717 & 0.654\\
     & 336 & 0.316 & 0.424 &  0.311 & 0.415 &0.119 & 0.202 & 0.264 & 0.366 & 5.805 &1.445 & 0.295 & 0.4396 & 0.260 & 0.364 & 0.142 & 0.243 & 0.756 & 0.670\\
     & 720 & 0.332 & 0.437 &0.327 & 0.428 & 0.135 & 0.224 & 0.305 & 0.400 &3.02 & 1.100 & 0.315 & 0.410 &0.289 & 0.387 & 0.153 & 0.251 & 0.799 & 0.688\\
     
     \label{tab: Appendix Univariate Forecasting}
 \end{tabular}
    \end{adjustbox}   
\end{table*}

\clearpage
\subsection{Robustness Testing} \label{appendix sub: Robustness Testing}

Table \ref{tab: Appendix Robustness Testing} presents a comparison of the HADL framework with baseline models, reported in terms of \ac{MSE} and \ac{MAE} across all benchmark datasets. Robustness testing was conducted by introducing noise into the training data ($X_{\text{train}}, y_{\text{train}}$) using the formula:
\begin{equation}
X_{\text{train}} := X_{\text{train}} + \mathcal{N}(0,1) * \eta 
\end{equation}
Here $ \mathcal{N}(0,1)$ represents random noise sampled from a standard normal distribution, and $\eta = \{0.0, 0.3, 0.7, 1.7, 2.3 \}$ controls the noise intensity. The validation and test datasets remained unchanged to ensure a consistent and unbiased evaluation of model robustness. All models are trained for 50 epochs with a patience level of 10.
\begin{table*}[htbp]
    \centering
    \caption{MSE and MAE results for robustness testing on benchmark datasets, comparing our HADL framework with (w/) and without (w/o) regularization against baseline models. The tests are conducted using a lookback window of $L=512$ and a prediction length of $H=192$ under varying noise intensities $\eta = \{0.0, 0.3, 0.7, 1.3, 1.7, 2.3\}$.}
    \begin{adjustbox}{width=0.9\textwidth}
    \renewcommand{\arraystretch}{1.5}
    \begin{tabular}{cc|cc|cc|cc|cc|cc|cc|cc|cc|cc}
    \multicolumn{2}{c|}{} & \multicolumn{10}{c|}{MLP} & \multicolumn{6}{c|}{Transformers} & \multicolumn{2}{c}{CNN} \\
    \cline{3-20}
     & & \multicolumn{2}{c|}{HADL (w/)} & \multicolumn{2}{c|}{HADL (w/o)} & \multicolumn{2}{c|}{DLinear (\citeyear{zeng2023transformers})} & \multicolumn{2}{c|}{SparseTSF (\citeyear{lin2024sparsetsf})} & \multicolumn{2}{c|}{FreTS (\citeyear{yi2024frequency})} & \multicolumn{2}{c|}{PatchTST (\citeyear{nie2022time})} & \multicolumn{2}{c|}{iTransformer (\citeyear{liu2023itransformer})} & \multicolumn{2}{c|}{FrNet (\citeyear{zhang2024frnet})} & \multicolumn{2}{c}{ModernTCN (\citeyear{luo2024moderntcn})}\\
     Dataset & H & MSE & MAE & MSE & MAE & MSE & MAE & MSE & MAE & MSE & MAE & MSE & MAE & MSE & MAE & MSE & MAE & MSE & MAE\\
     \hline
     \multirow{5}{*}{ETTh1}& 0.0 & 0.396 & 0.414 & 0.397 & 0.414 & 0.422 & 0.429 & 0.412 & 0.421 & 0.517 & 0.503 & 0.427 & 0.442 & 0.427 & 0.443 & 0.396 & 0.416 & 0.421 & 0.430 \\
     
     &0.3 & 0.396 & 0.414 & 0.397 & 0.414 & 0.415 & 0.424 & 0.412 & 0.422 & 0.516 & 0.504 & 0.428 & 0.443 & 0.434 & 0.448 & 0.394 & 0.414 & 0.413 & 0.425 \\
     
     &0.7 & 0.398 & 0.417 & 0.398 & 0.417 & 0.411 & 0.423 & 0.420 & 0.430 & 0.521 & 0.506 & 0.430 & 0.445 & 0.434 & 0.447 & 0.395 & 0.415 & 0.408 & 0.426 \\
     
     &1.3 & 0.406 & 0.424 & 0.404 & 0.423 & 0.412 & 0.428  & 0.436 & 0.441 & 0.528 & 0.510 & 0.439 & 0.453 & 0.440 & 0.449 & 0.406 & 0.425 & 0.409 & 0.426 \\
     
     &1.7 & 0.413 & 0.429 & 0.411 & 0.427 & 0.416 & 0.433& 0.440 & 0.443 & 0.544 & 0.519 & 0.446 & 0.458 & 0.470 & 0.463 & 0.410 & 0.427 & 0.413 & 0.429 \\
     
     &2.3 & 0.427 & 0.437 & 0.423 & 0.435 & 0.424 & 0.441 & 0.454 & 0.451 & 0.575 & 0.536 & 0.459 & 0.466 & 0.525 & 0.490 & 0.423 & 0.435 & 0.422 & 0.435 \\
     \hline
     
     \multirow{5}{*}{ETTh2} & 0.0 & 0.333 & 0.377 & 0.334 & 0.377 & 0.547  & 0.490 & 0.343 & 0.381 & 0.581 & 0.534 & 0.348 & 0.396 & 0.383 & 0.409 & 0.329 & 0.379 & 0.337 & 0.387 \\
     
     &0.3 & 0.333 & 0.377 & 0.333 & 0.377 & 0.540 & 0.483 & 0.348 & 0.382 & 0.541 & 0.510 & 0.351 & 0.398 & 0.379 & 0.404 & 0.331 & 0.377 & 0.332 & 0.381 \\
     
     &0.7 & 0.334 & 0.379 & 0.334 & 0.378 & 0.541 & 0.485 & 0.351 & 0.385 & 0.549 & 0.517 & 0.352 & 0.398 & 0.370 & 0.400 & 0.330 & 0.376 & 0.335 & 0.380 \\
     
     &1.3 & 0.338 & 0.382 & 0.338 & 0.381 & 0.554 & 0.494 & 0.354 & 0.390 & 0.522 & 0.505 & 0.352 & 0.398 & 0.379 & 0.403 & 0.337 & 0.379 & 0.340 & 0.382 \\
     
     &1.7 & 0.341 & 0.385 & 0.341 & 0.384 & 0.566 & 0.501 & 0.362 & 0.396 & 0.576 & 0.531 & 0.343 & 0.390 & 0.380 & 0.406 & 0.340 & 0.382 & 0.335 & 0.380 \\
     
     &2.3 & 0.346 & 0.390 & 0.345 & 0.388 & 0.587 & 0.513 & 0.368 & 0.401 & 0.524 & 0.510 & 0.343 & 0.391 & 0.386 & 0.414 & 0.345 & 0.386 & 0.346 & 0.387 \\
     \hline
     
     \multirow{5}{*}{ETTm1} & 0.0 & 0.336 & 0.366 & 0.338 & 0.367 & 0.333 & 0.362 & 0.355 & 0.377 & 0.446 & 0.459 & 0.335  & 0.367 & 0.356 & 0.394 & 0.336 & 0.368 & 0.369 & 0.396 \\
     
     &0.3 & 0.337 & 0.367 & 0.339 & 0.367 & 0.334 & 0.362 & 0.354 & 0.377 & 0.457 & 0.468 & 0.337  & 0.368 & 0.350 & 0.390 & 0.342 & 0.373 & 0.369 & 0.393 \\
     
     &0.7 & 0.341 & 0.372 & 0.342 & 0.371 & 0.337 & 0.366  & 0.354 & 0.378 & 0.452 & 0.461 & 0.344 & 0.374 & 0.354 & 0.391 & 0.344 & 0.375 & 0.372 & 0.395 \\
     
     &1.3 & 0.350 & 0.380 & 0.349 & 0.379 & 0.343 & 0.374  & 0.355 & 0.383 & 0.443 & 0.453 & 0.350 & 0.380 & 0.374 & 0.405 & 0.347 & 0.379 & 0.365 & 0.392 \\
     
     &1.7 & 0.356 & 0.386 & 0.355 & 0.384 & 0.349 & 0.380 & 0.359 & 0.387 & 0.449 & 0.457 & 0.352 & 0.382 & 0.389 & 0.414 & 0.355 & 0.385 & 0.366 & 0.394 \\
     
     &2.3 & 0.369 & 0.395 & 0.366 & 0.393 & 0.358 & 0.389 & 0.371 & 0.396 & 0.452 & 0.459 & 0.360 & 0.390 & 0.408 & 0.422 & 0.359 & 0.394 & 0.370 & 0.400 \\
     \hline
     
     \multirow{5}{*}{ETTm2}&  0.0 & 0.218 & 0.290 & 0.218 & 0.290 & 0.302 & 0.355 & 0.225 & 0.296 & 0.347 & 0.406 & 0.223 & 0.295 & 0.243 & 0.312 & 0.219 & 0.292 & 0.241 & 0.313 \\
     
     &0.3 & 0.218 & 0.290 & 0.218 & 0.290 & 0.304 & 0.356 & 0.224 & 0.295 & 0.332 & 0.398 & 0.223 & 0.295 & 0.240 & 0.309 & 0.219 & 0.292 & 0.235 & 0.307 \\
     
     &0.7 & 0.220 & 0.293 & 0.220 & 0.292 & 0.312 & 0.363 & 0.225 & 0.296 & 0.323 & 0.390 & 0.224 & 0.297 & 0.248 & 0.314 & 0.220 & 0.293 & 0.232 & 0.303 \\
     
     &1.3 & 0.226 & 0.298 & 0.225 & 0.297 & 0.328 & 0.376 & 0.230 & 0.300 & 0.309 & 0.382 & 0.227 & 0.297 & 0.261 & 0.323 & 0.230 & 0.302 & 0.233 & 0.0.304 \\
     
     &1.7 & 0.231 & 0.303 & 0.229 & 0.301 & 0.341 & 0.386 & 0.233 & 0.304 & 0.372 & 0.419 & 0.231 & 0.302 & 0.267 & 0.329 & 0.239 & 0.309 & 0.236 & 0.307 \\
     
     &2.3 & 0.238 & 0.309 & 0.236 & 0.308 & 0.363 & 0.402 & 0.239 & 0.310 & 0.335 & 0.401 & 0.234 & 0.304 & 0.266 & 0.334 & 0.239 & 0.309 & 0.243 & 0.313 \\
     \hline
    
    \multirow{5}{*}{Traffic}&  0.0 & 0.433 & 0.311 & 0.433 & 0.308 & 0.430 & 0.303 & 0.441 & 0.309 & 0.417 & 0.286 & 0.498 & 0.357& 0.462 & 0.343  & 0.387 & 0.283  & 0.431 & 0.304\\
    
     &0.3 & 0.434 & 0.311 & 0.434 & 0.308 & 0.423 & 0.299 & 0.448 & 0.329 & 0.414 & 0.287 & 0.500 & 0.357 & 0.467 & 0.345 & 0.387 & 0.283 & 0.437 & 0.309 \\
     
     &0.7 & 0.442 & 0.315 & 0.440 & 0.310 & 0.433 & 0.305 & 0.475 & 0.325 & 0.460 & 0.298 & 0.509 & 0.361 & 0.514 & 0.382 & 0.414 & 0.317 & 0.460 & 0.324 \\
     
     &1.3 & 0.469 & 0.332 & 0.464 & 0.325 & 0.468 & 0.324 & 0.517 & 0.354 & 0.463 & 0.317 & 0.539 & 0.376 & 0.673 & 0.482 & 0.461 & 0.361 &  0.494 & 0.382\\
     
     &1.7 & 0.494 & 0.347 & 0.487 & 0.338 & 0.495 & 0.338 & 0.551 & 0.376 & 0.479 & 0.326 & 0.565 & 0.389 & 0.782 & 0.537 & 0.480 & 0.374 &  0.545 & 0.374\\
     
     &2.3 & 0.534 & 0.370 & 0.525 & 0.361 & 0.537 & 0.359 & 0.623 & 0.432 & 0.505 & 0.338 & 0.601 & 0.407 & 0.716 & 0.518 & 0.482 & 0.380 &  0.596 & 0.403\\
     \hline

    \multirow{5}{*}{Weather}&  0.0 & 0.214 & 0.260 & 0.214 & 0.260 & 0.184 & 0.249 & 0.222 & 0.268 & 0.222 & 0.290 & 0.189 & 0.239 & 0.202 & 0.249 & 0.186 & 0.236 & 0.196 & 0.246\\
    
     &0.3 & 0.215 & 0.262 & 0.214 & 0.260 & 0.189 & 0.255 & 0.222 & 0.268 & 0.225 & 0.294 & 0.191 & 0.243 & 0.208 & 0.253 & 0.200 & 0.246 & 0.206 & 0.251 \\
     
     &0.7 & 0.217 & 0.265 & 0.216 & 0.263 & 0.195 & 0.262 & 0.222 & 0.268 & 0.235 & 0.305 & 0.197 & 0.249 & 0.219 & 0.264 & 0.224 & 0.268 & 0.212 & 0.256 \\
     
     &1.3 & 0.222 & 0.270 & 0.221 & 0.269 & 0.202 & 0.271 & 0.225 & 0.272 & 0.230 & 0.293 & 0.204 & 0.256 & 0.227 & 0.275 & 0.226 & 0.273 & 0.215 & 0.261 \\
     
     &1.7 & 0.225 & 0.274 & 0.224 & 0.273 & 0.207 & 0.277 & 0.228 & 0.275 & 0.229 & 0.291 & 0.208 & 0.259 & 0.234 & 0.280 & 0.227 & 0.275 & 0.213 & 0.262 \\
     
     &2.3 & 0.230 & 0.280 & 0.229 & 0.278 & 0.214 & 0.287 & 0.231 & 0.280 & 0.230 & 0.290 & 0.215 & 0.266 & 0.243 & 0.288 & 0.231 & 0.279 &  0.218 & 0.269\\
     \hline

     \multirow{5}{*}{Electricity}&  0.0 & 0.168 & 0.271 & 0.168 & 0.271 & 0.148 & 0.245  & 0.164 & 0.259 & 0.215 & 0.331 & 0.172 & 0.277 & 0.168 & 0.270 & 0.146 & 0.239 & 0.175 & 0.279 \\
    
     &0.3 & 0.169 & 0.272 & 0.168 & 0.272 & 0.148 & 0.246 & 0.168 & 0.266 & 0.219 & 0.334 & 0.174 & 0.279 & 0.174 & 0.277 & 0.156 & 0.254 & 0.174 & 0.279 \\
     
     &0.7 & 0.175 & 0.279 & 0.174 & 0.277 & 0.154 & 0.256 & 0.188 & 0.288 & 0.219 & 0.334 & 0.180 & 0.286 & 0.206 & 0.318 & 0.167 & 0.269 & 0.178 & 0.285  \\
     
     &1.3 & 0.189 & 0.294 & 0.189 & 0.293 & 0.167 & 0.274 & 0.212 & 0.311 & 0.233 & 0.346 & 0.193 & 0.301 & 0.313 & 0.426 & 0.191 & 0.298 & 0.184 & 0.293 \\
     
     &1.7 & 0.200 & 0.305 & 0.199 & 0.304 & 0.176 & 0.285 & 0.229 & 0.328 & 0.245 & 0.355 & 0.202  & 0.310 & 0.351 & 0.456 & 0.207 & 0.317 & 0.188 & 0.297 \\
     
     &2.3 & 0.217 & 0.321 & 0.216 & 0.320 & 0.190 & 0.301 & 0.258 & 0.345 & 0.253 & 0.362  & 0.218 & 0.327 & 0.424 & 0.510  & 0.233 & 0.344 & 0.194 & 0.303 \\
    \end{tabular}
     \label{tab: Appendix Robustness Testing}
    \end{adjustbox} 
\end{table*}

\end{document}